\documentclass[journal]{IEEEtran} 
\usepackage{cite}
\usepackage{amsmath,amssymb,amsfonts}
\usepackage{graphicx}
\usepackage{array}
\usepackage{units}
\usepackage{makecell}
\usepackage{multirow}
\usepackage[nolist,nohyperlinks]{acronym}
\usepackage{textcomp}
\usepackage{breqn}
\usepackage{mathtools}
\usepackage{verbatim}
\usepackage[export]{adjustbox}
\usepackage{subfig}
\usepackage{mwe}    
\usepackage{tikz}

\begin{document}

\title{VersaVIS: An Open Versatile Multi-Camera Visual-Inertial Sensor Suite}
\author{Florian Tschopp$^1$, Michael Riner$^1$, Marius Fehr$^{1,2}$, Lukas Bernreiter$^1$, 
Fadri Furrer$^1$, Tonci Novkovic$^1$, Andreas Pfrunder$^3$, Cesar Cadena$^1$, Roland Siegwart$^1$, and Juan Nieto$^1$%
\thanks{$^1$Authors are members of the Autonomous Systems Lab, ETH Zurich, Switzerland; {\tt\small \{firstname.lastname\}@mavt.ethz.ch}}%
\thanks{$^2$Marius Fehr is with Voliro Airborne Robotics, Switzerland; {\tt\small marius.fehr@voliro.ch}}%
\thanks{$^3$Andreas Pfrunder is with Sevensense Robotics, Switzerland; {\tt\small andreas.pfrunder@sevensense.ch}}%
}

\maketitle

\begin{abstract}
Robust and accurate pose estimation is crucial for many applications in mobile robotics. Extending visual \ac{slam} with other modalities such as an \ac{imu} can boost robustness and accuracy. However, for a tight sensor fusion, accurate time synchronization of the sensors is often crucial. Changing exposure times, internal sensor filtering, multiple clock sources and unpredictable delays from operation system scheduling and data transfer can make sensor synchronization challenging. In this paper, we present VersaVIS, an Open Versatile Multi-Camera Visual-Inertial Sensor Suite aimed to be an efficient research platform for easy deployment, integration and extension for many mobile robotic applications. 
VersaVIS provides a complete, open-source hardware, firmware and software bundle to perform time synchronization of multiple cameras with an \ac{imu} featuring exposure compensation, host clock translation and independent and stereo camera triggering. The sensor suite supports a wide range of cameras and \acp{imu} to match the requirements of the application.
The synchronization accuracy of the framework is evaluated on multiple experiments achieving timing accuracy of less than $\unit[1]{ms}$. Furthermore, the applicability and versatility of the sensor suite is demonstrated in multiple applications including visual-inertial \ac{slam}, multi-camera applications, multi-modal mapping, reconstruction and object based mapping.
\end{abstract}

\section{Introduction}
Autonomous mobile robots are well established in controlled environments such as factories where they rely on external infrastructure such as magnetic tape on the floor or beacons. However, in unstructured and changing environments, robots need to be able to plan their way and interact with their environment which, as a first step, requires accurate positioning \cite{RolandSiegwart2011}.\\

In mobile robotic applications, visual sensors can provide solutions for odometry and \acf{slam}, achieving good accuracy and robustness. Using additional sensor modalities such as \acfp{imu} \cite{Bloesch2015,Leutenegger2015,Qin2018VINS-Mono:Estimator} can additionally improve robustness and accuracy for a wide range of applications. For many frameworks, a time offset or delay between modalities can lead to bad results or even render the whole approach unusable. There are frameworks such as VINS-Mono \cite{Qin2018VINS-Mono:Estimator} that can estimate a time offset during estimation, however, convergence of the estimation can be improved by accurate timestamping, i.e. assigning timestamps to the sensor data that precisely correspond to measurement time from a common clock across all sensors.\\
For a reliable and accurate sensor fusion, all sensors need to provide  timestamps matchable between sensor modalities.
\begin{figure}[t!]
\centering
    \subfloat[Lidarstick] {%
        \includegraphics[width=0.45\columnwidth, trim={4cm 2.5cm 4cm 12.5cm}, clip, valign=m]{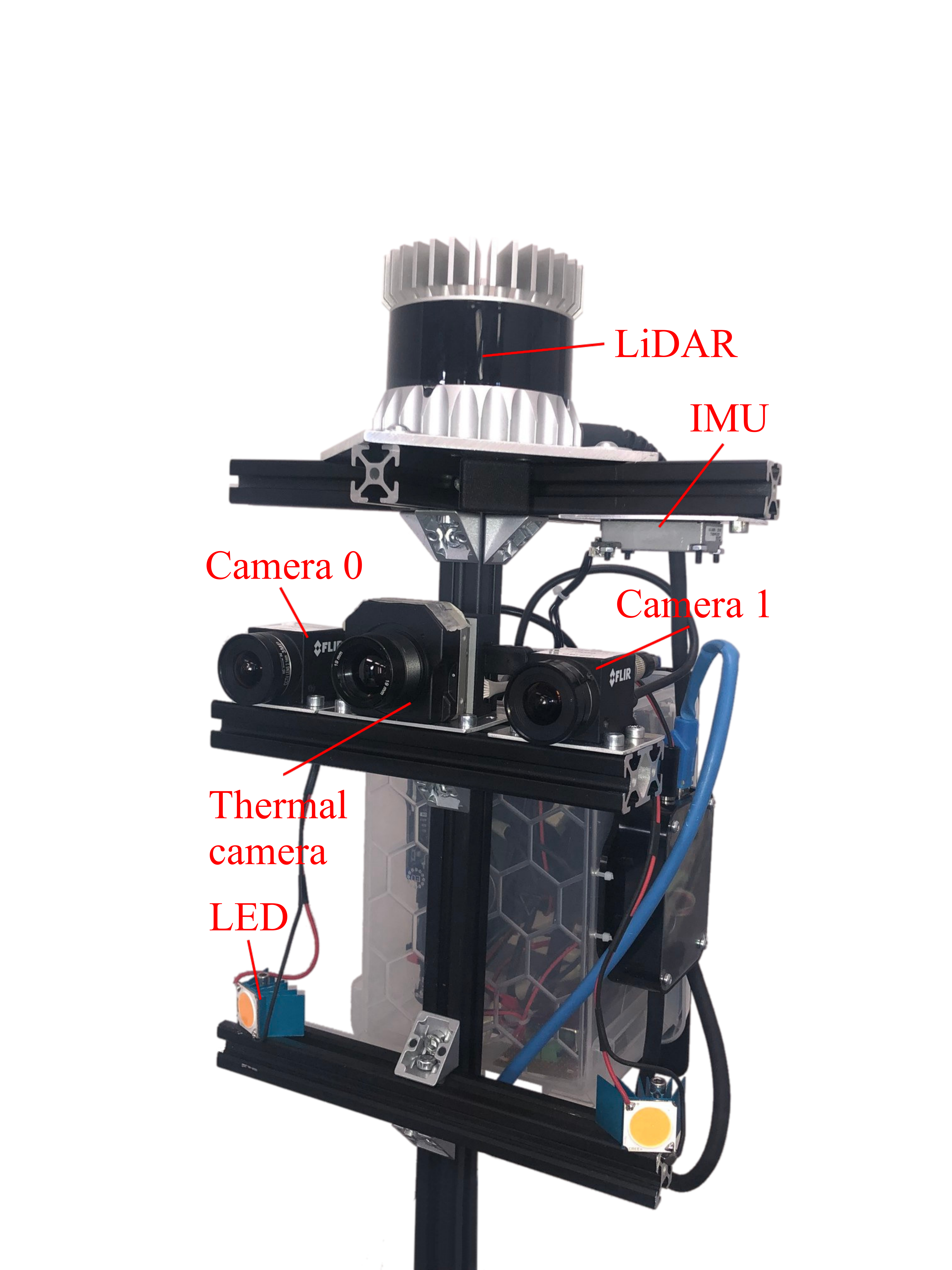}%
        \vphantom{\includegraphics[width=0.45\columnwidth, trim={0cm 2.2cm 0cm 0cm}, clip,valign=m]{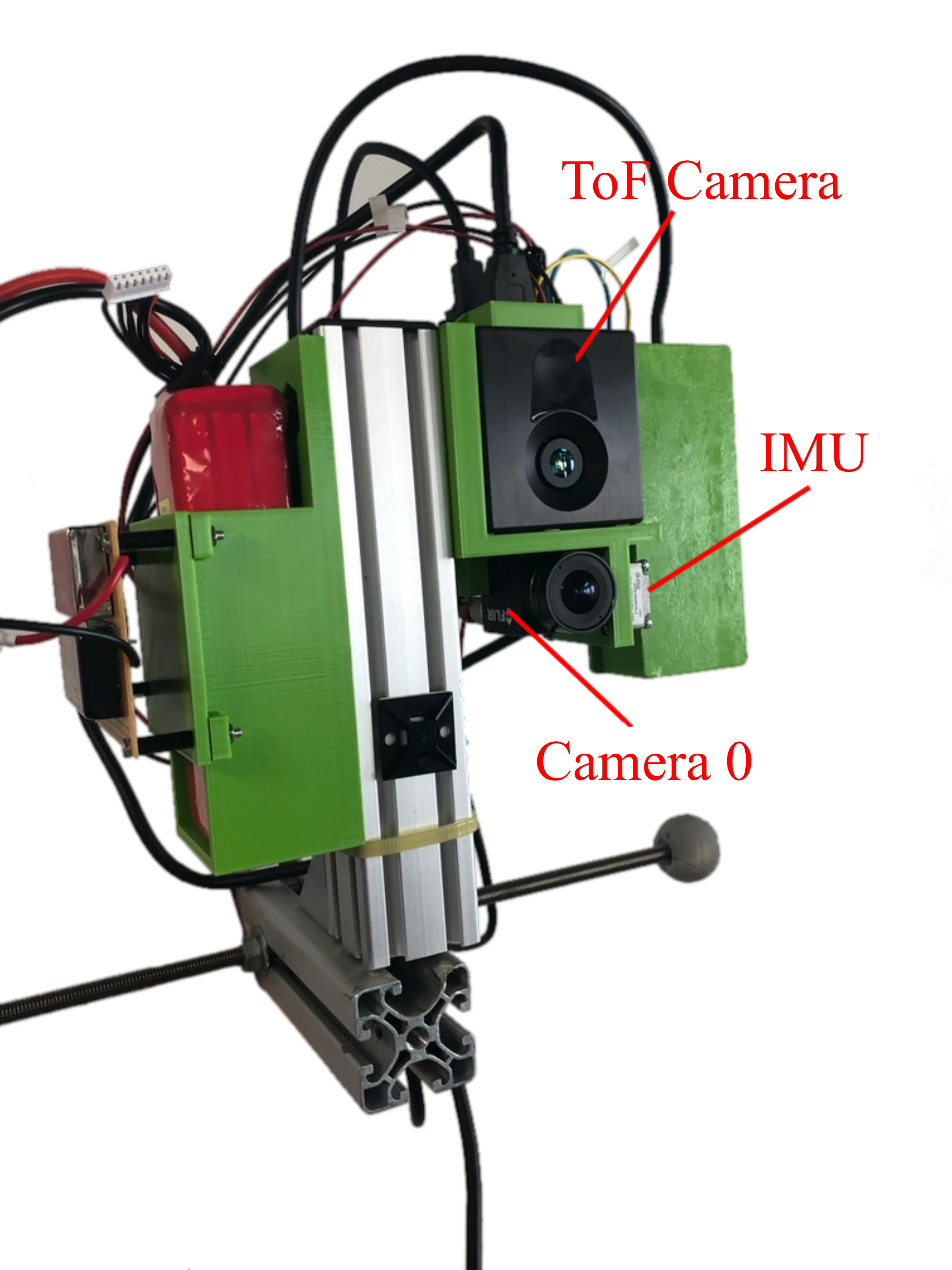}}%
        \label{fig:lidarstick}
    } \hfill
    \subfloat[\ac{rgbdi} sensor] {%
        \includegraphics[width=0.45\columnwidth, trim={0cm 2.2cm 0cm 0cm}, clip, valign=m]{figures/rgbdi2_resized_with_markers.pdf}
        \label{fig:rgbdi}
    } \\
    \subfloat[Stereo \ac{vi} Sensor \cite{Tschopp2019ExperimentalVehiclesb}] {%
        \includegraphics[width=0.45\columnwidth, valign=m]{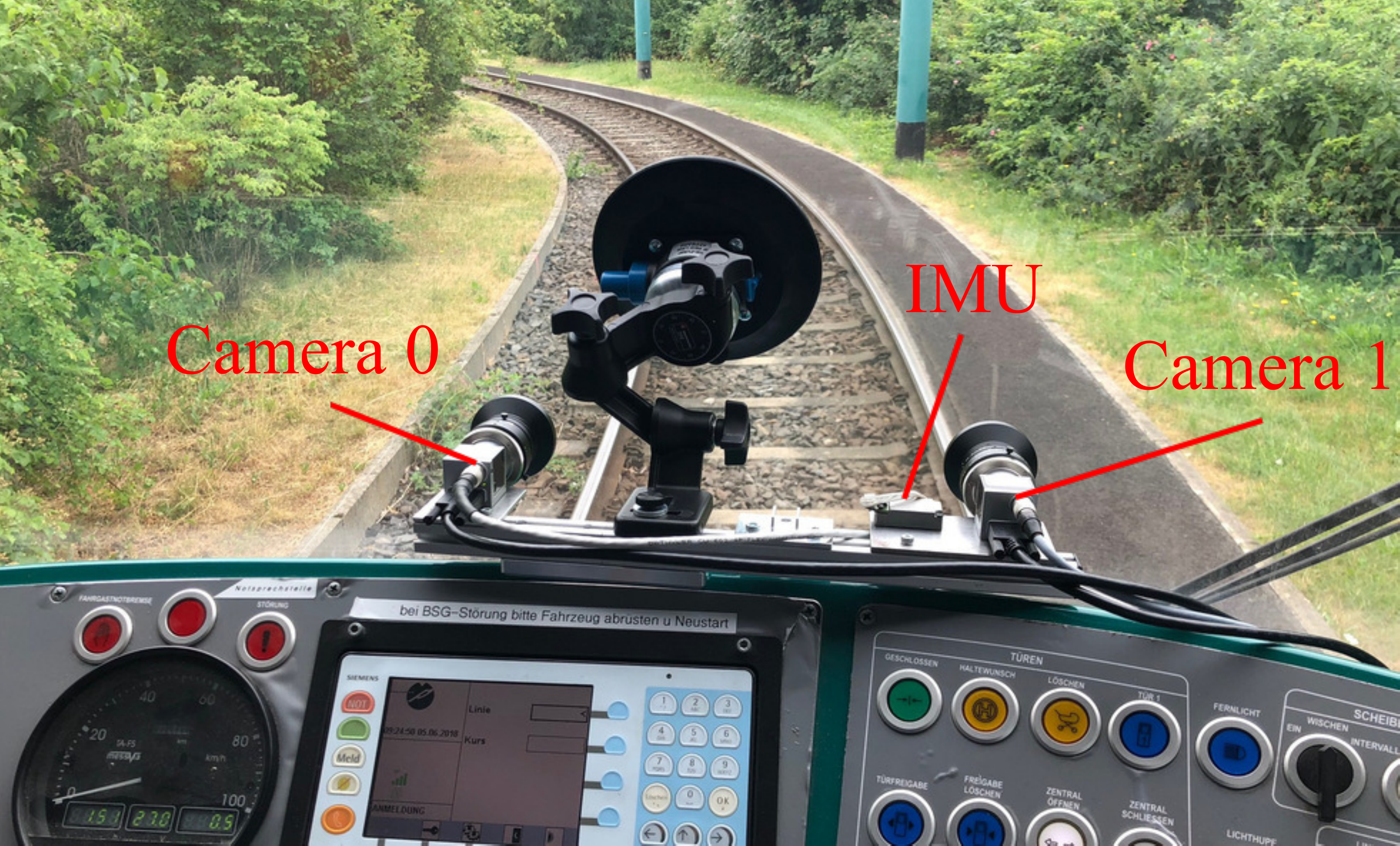}
        \vphantom{\includegraphics[width=0.45\columnwidth, valign=m]{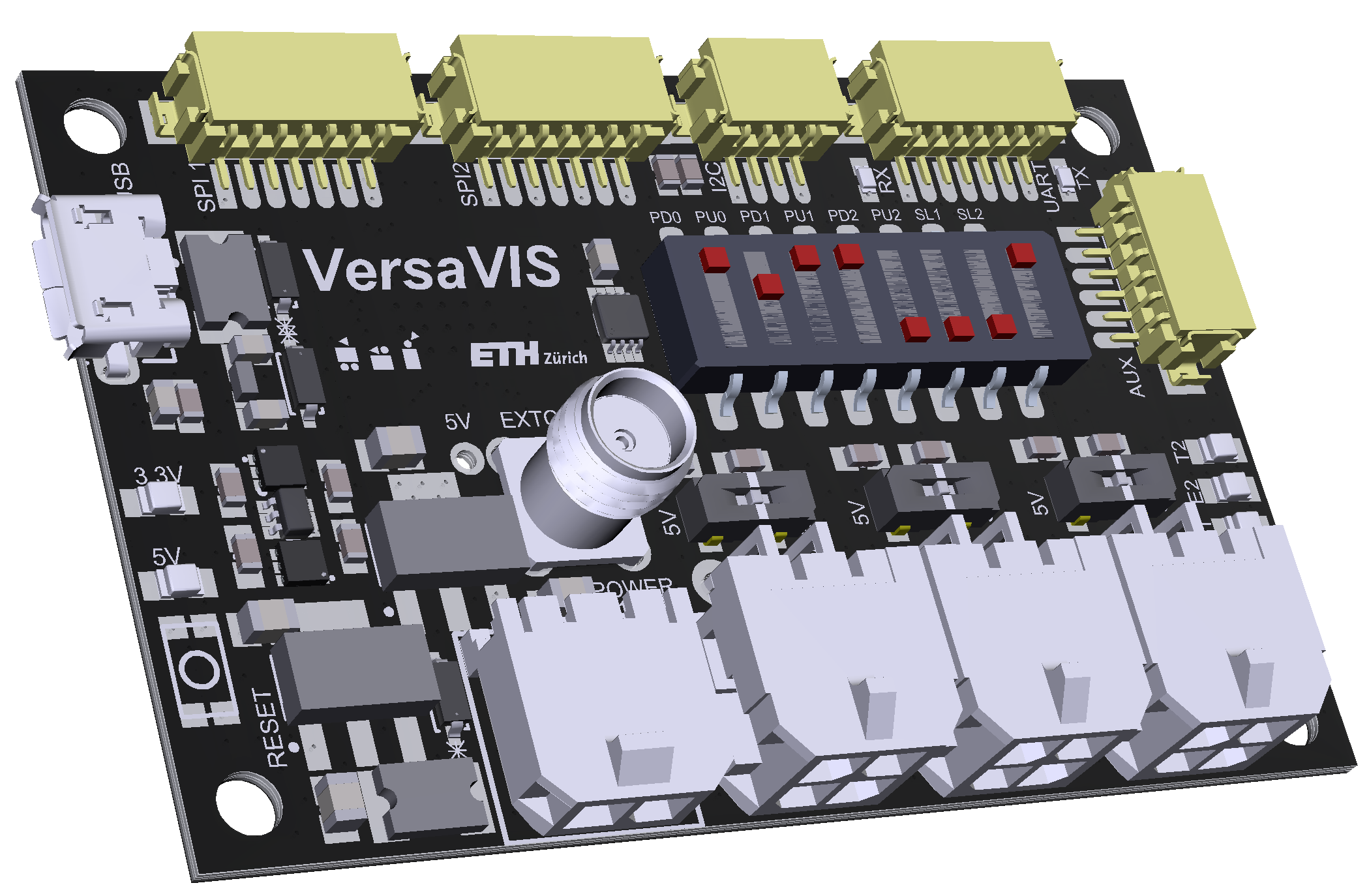}}%
        \label{fig:stereo}
    } \quad
    \subfloat[\acs{VersaVIS} triggering board] {%
        \includegraphics[width=0.45\columnwidth, valign=m]{figures/pcb.png}
        \label{fig:versavis}
    }
    \caption{The \acs{VersaVIS} sensor in different configurations. \acs{VersaVIS} is able to synchronize multiple sensor modalities such as \acp{imu} and cameras (e.g. monochrome, color, \ac{tof} and thermal) but can also be used in conjunction with additional sensors such as \acp{lidar}.}
    \label{fig:sensors}
\end{figure}
Typically, the readout of camera image and \ac{imu} measurement are not on the same device resulting in a clock offset between the two timestamps which is hard to predict due to \ac{usb} buffer delay, \ac{os} scheduling, changing exposure times and internal sensor filtering. Therefore, measurement correspondences between modalities are ambiguous and assignment on the host is not trivial. \\
Currently, there is no reference sensor synchronization framework for data collection, which makes it hard to compare results obtained in various publications that deal with \acf{vi} \ac{slam}. Commercially and academically established sensors such as the Skybotix \ac{vi}-Sensor \cite{Nikolic2014}, Intel RealSense \cite{IntelCorporation2019IntelT265}, SkyAware sensor based on work from Honegger~et.~al.~\cite{Honegger2017EmbeddedStereo} or PIRVS \cite{ZheZhang2018} are either unavailable and/or are limited in  hardware configuration regarding image sensor, lens, camera baseline and \ac{imu}. Furthermore, it is often impossible to add further extensions of those frameworks to enable fusion with other modalities such as \ac{lidar} sensors or external illumination.\\

In this paper, we introduce \acs{VersaVIS}\footnote{\acs{VersaVIS} is available at: www.github.com/ethz-asl/versavis} shown in Figure~\ref{fig:sensors}, the first open-source framework that is able to accurately synchronize a large range of camera and \ac{imu} sensors. The sensor suite is aimed for the research community to enable rapid prototyping of affordable sensor setups in different fields in mobile robotics where visual navigation is important. Special emphasis is put on an easy integration for different applications and easy extensibility by being based on well-known open-source frameworks such as Arduino~\cite{ArduinoInc.2019ArduinoZero} and \ac{ros}\cite{StanfordArtificialIntelligenceLaboratoryetal.2018RoboticSystem}.\\
The remainder of the paper is organized as follows: In Section~\ref{sec:versavis}, the sensor suite is described in detail including all of its features. Section~\ref{sec:evaluations} provides evaluations of the synchronization accuracy of the proposed framework. Finally, Section~\ref{sec:applications} showcases the use of the \ac{VersaVIS} in multiple applications while Section~\ref{sec:conclusions} provides a conclusion with an outlook on future work.

\section{The Visual-Inertial Sensor Suite} \label{sec:versavis}
\begin{figure}
    \centering
    \includegraphics[width=\columnwidth]{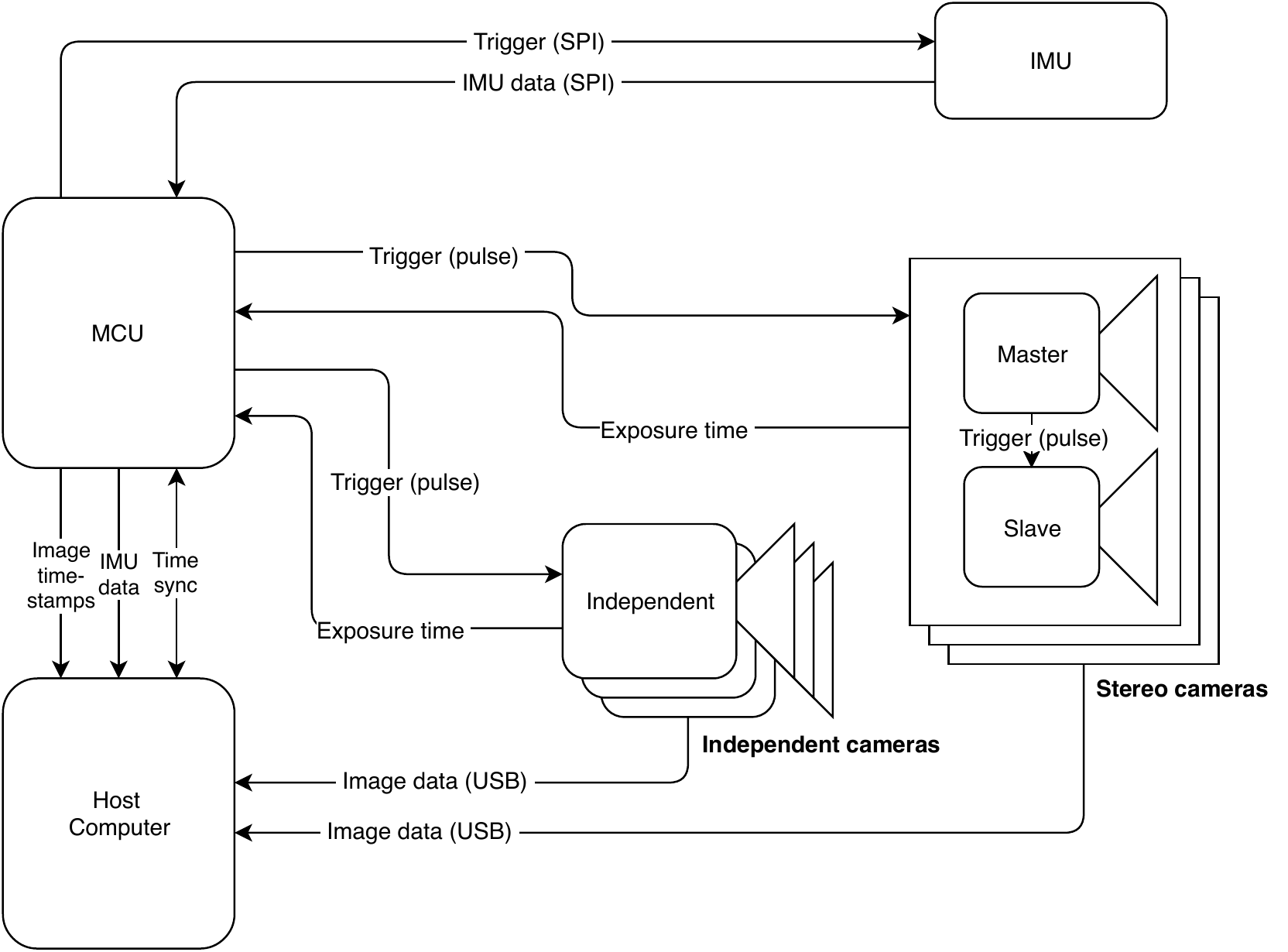}    
    \caption{Design overview of \ac{VersaVIS}. The \ac{mcu} sends triggers to both \ac{imu} and the connected cameras. Image data is directly transferred to the host computer where it is combined with the timestamps from the \ac{mcu}.}
    \label{fig:design_overview}
\end{figure}

The proposed sensor suite consists of three different parts, (i) the firmware which runs on the \ac{mcu}, (ii) the host driver running on a \ac{ros} enabled machine, and (iii) the hardware trigger \ac{pcb}. An overview of the framework is provided in Figure~\ref{fig:design_overview}. Here, the procedure is described for a reference setup consisting of two cameras and an \ac{spi} enabled \ac{imu}.\\
The core component of \ac{VersaVIS} is the \ac{mcu}. First of all, it is used to periodically trigger the \ac{imu} readout together with setting the timestamps and sending the data to the host. Furthermore, the \ac{mcu} sends triggering pulses to the cameras to start image exposure. This holds for both independent cameras and stereo cameras (see Section~\ref{sec:firmware}). After successful exposure, the \ac{mcu} reads the exposure time by listening to the cameras' exposure signal in order to perform exposure compensation described in Section~\ref{sec:std_cameras} and setting mid-exposure timestamps. The timestamps are sent to the host together with a strictly increasing sequence number. The image data is hereby sent directly from the camera to the host computer to avoid massive amounts of data through the \ac{mcu}. This enables to use high-resolution cameras even with a low-performance \ac{mcu}. \\
Finally, the host computer merges image timestamps from the \ac{mcu} with the corresponding image messages based on a sequence number (see Section~\ref{sec:std_cameras}).
\subsection{Firmware}\label{sec:firmware}
The \ac{mcu} is responsible for triggering the devices at the correct time and to capture timestamps of the triggered sensor measurements. This is based on the usage of hardware timers and external signal interrupts.
\subsubsection{Standard cameras} \label{sec:std_cameras}
\begin{figure}
\centering
\includegraphics[width=\columnwidth]{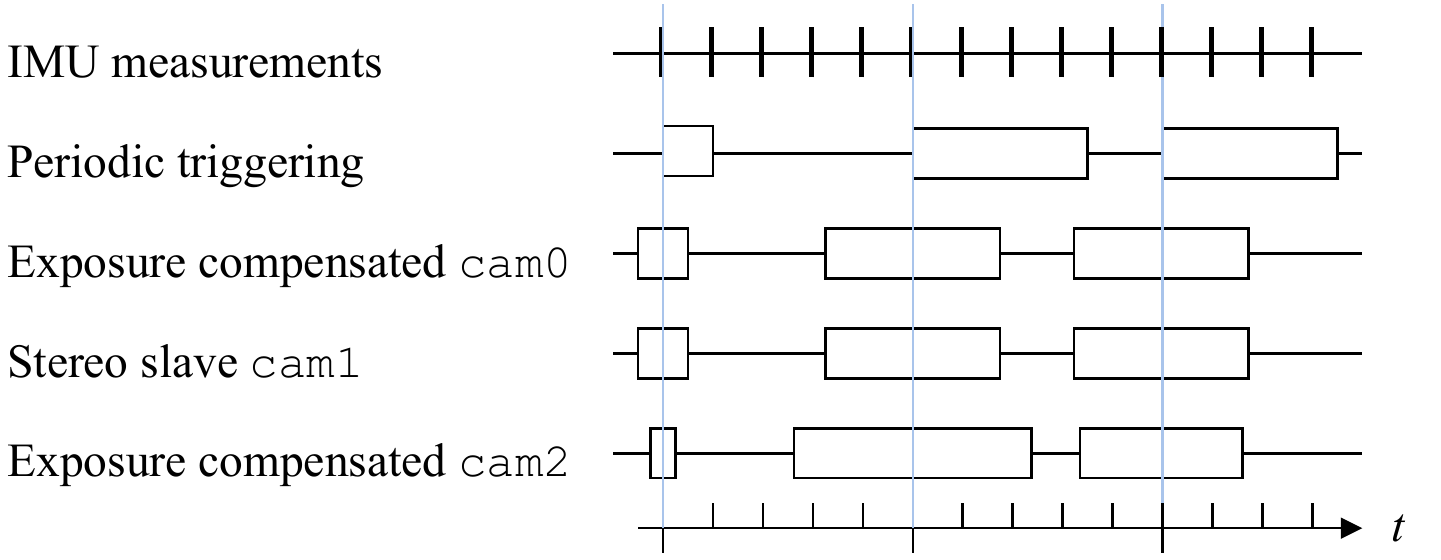}
\caption{Exposure time compensation for multi-camera \ac{vi} sensor setup (adapted from \cite{Nikolic2014}). The blue lines indicate corresponding measurements.} \label{fig:exp_comp}
\end{figure}

In the scope of the \ac{mcu}, standard cameras are considered sensors that are triggerable with signal pulses and with non-zero data measurement time, i.e. image exposure time. Furthermore, the sensors need to provide an exposure signal (often called strobe signal) which indicates the exposure state of the sensor. While the trigger pulse and the timestamp are both created on the \ac{mcu} based on its internal clock, the data is transferred via \ac{usb} or Ethernet directly to the host computer. To enable correct association of the image data and timestamp on the host computer (see Section~\ref{sec:synchronizer}), both the timestamp and the image data are assigned an independent sequence number $n_{VersaVIS}$ and $n_{image}$, respectively. The mapping between these sequence numbers is determined during initialization as a simultaneous start of the cameras and the trigger board cannot be guaranteed.
\begin{itemize}
    \item Initialization procedure: 
        After startup of the camera and trigger board, corresponding sequence numbers are found by very slowly trigger the camera without exposure time compensation. Corresponding sequence numbers are then determined by closest timestamps. This holds true if the triggering period time is significantly longer than the expected delay and jitter on the host computer. As soon as the sequence number offset $o_{cam}$ is determined, exposure compensation mode at full frame rate can be used.
    \item Exposure time compensation:
        Performing \ac{ae}, the camera adapts its exposure time to the current illumination resulting in a non-constant exposure time. Furgale et al. \cite{Furgale2013UnifiedSystems} showed, that mid-exposure timestamping is beneficial for image based state estimation, especially when using global shutter cameras. Instead of periodically triggering the camera, a scheme proposed by Nikolic~et~al.~\cite{Nikolic2014} is employed. The idea is to trigger the camera for a periodic mid-exposure timestamp by starting exposure half the exposure time earlier to its mid-exposure timestamp as shown in Figure~\ref{fig:exp_comp} for \texttt{cam0}, \texttt{cam1} and \texttt{cam2}. The exposure time return signal is used to time the current exposure time and calculate the offset to the mid-exposure timestamp of the next image. Using this approach, corresponding measurements can be obtained even if multiple cameras do not share the same exposure time (e.g. \texttt{cam0} and \texttt{cam2} in Figure~\ref{fig:exp_comp}). 
    \item Master-slave mode:
        Using two cameras in a stereo setup compared to a monocular camera can retrieve metric scale by stereo matching. This can enable certain applications where \ac{imu} excitation is not high enough  and therefore biases are not fully observable without this scale input e.g. for rail vehicles described in Section~\ref{sec:trains}. Furthermore, it can also provide more robustness. To perform accurate and efficient stereo matching, it is highly beneficial if keypoints from the same spot have a similar appearance. This can be achieved by using the exact same exposure time on both cameras. Thereby, one camera serves as the master performing \ac{ae} while the other adapts its exposure time. This is achieved by routing the exposure signal from \texttt{cam0} directly to the trigger of \texttt{cam1} and also using it to determine the exposure time for compensation.
\end{itemize}
\subsubsection{Other triggerable sensors}
Some sensors enable measurement triggering but do not require or offer the possibility to do exposure compensation e.g. thermal cameras or \ac{tof} cameras. These typically do not allow for adaptive exposure compensation but rather have a fixed exposure/integration time. They can be treated the same as a standard camera, but with fixed exposure time. \\
Sensors that provide immediate measurements (such as external \acp{imu}) do not need an exposure time compensation and can just use a standard timer. Note that the timestamp for those sensors are still captured on the \ac{mcu} and therefore correspond to the other sensor modalities.
\subsubsection{Other non-triggerable sensors} \label{sec:add_sensors}
In robotics, it is often useful to perform sensor fusion with multiple available sensor modalities such as wheel odometers or \ac{lidar} sensors \cite{Zhang2015Visual-lidarFast}. Most of such sensor hardware do not allow triggering or low-level sensor readout but send the data continuously to a host computer. In order to enable precise and accurate sensor fusion, having corresponding timestamps of all sensor modalities is often crucial. \\
As most such additional sensors produce their timestamps based on the host clock or synchronized to the host clock, \ac{VersaVIS} performs time \textit{translation}\footnote{In this context, clock \textit{synchronization} refers to modifying the slave clock speed to align to the master clock while clock \textit{translation} refers to translating the timestamp of the slave clock to the time of the master clock \cite{Sommer2017ACameras}.} to the host. An on-board \ac{ekf} is deployed to estimate clock skew $\eta$ and clock offset $\delta$ using 
\begin{equation}
    t_{host} = t_{slave} + dt\cdot \eta^k + \delta^k + \Delta,
\end{equation}
where $t_{host}$ and $t_{slave}$ are the timestamps on the host and slave (in this case the \ac{VersaVIS} \ac{mcu}) respectively, $k$ is the update step, $dt =  t_{slave} - t_{slave}^{k}$ is the time since the last \ac{ekf} update and $\Delta$ refers to the initial clock offset set at the initial connection between host and slave.\\
Periodically, \ac{VersaVIS} performs a filter update by requesting the current time of the host \begin{equation}
    t_{host}^k = t_{host}^{a} - \frac{1}{2}\left( t_{slave}^{a} - t_{slave}^{r}\right),
\end{equation}
where $t^r$ is the time at sending the request and $t^a$ is the time when receiving the answer from the host assuming that communication time delay between host and slave is symmetric.
The filter update is then performed using standard \ac{ekf} equations:
\begin{equation} \label{eq:ekf_states}
\begin{split}
    \hat{\delta}^k& = \delta^{k-1} + dt \cdot \eta^k,\\
    \hat{\eta}^k &= \eta^{k-1},\\
    \hat{\mathsf{P}}^k &= \begin{bmatrix} 
1 &  dt\\
0 & 1 
\end{bmatrix} \cdot \mathsf{P}^{k-1} \cdot \begin{bmatrix} 
1 & dt\\
0 & 1 
\end{bmatrix}^{\top} + \begin{bmatrix} 
Q_{\delta} &  0\\
0 & Q_{\eta}
\end{bmatrix}
    \end{split}
\end{equation}
where $\hat{\cdot}$ depicts the prediction, $\mathsf{P}$ is the covariance matrix and $Q_{\delta, \eta}$ are the noise parameters of the clock offset and clock skew, respectively. The measurement residual $\epsilon$ can be written as
\begin{equation}
    \begin{split}
        \epsilon^k &= t_{host}^k - \Delta -
                         t_{slave}^k -
                        \hat{\delta}^k,
    \end{split}
\end{equation}
where the Kalman gain $\mathsf{K}$ and the measurement update can be derived using the standard \ac{ekf} equations.
\subsection{Host computer driver}
The host computer needs to run a lightweight application in order to make sure the data from \ac{VersaVIS} can be correctly used in the \ac{ros} environment.
\subsubsection{Synchronizer} \label{sec:synchronizer}
The host computer needs to take care of merging together the image data directly from the camera sensors and the image timestamps from the \ac{VersaVIS} triggering board. \\
During initialization, timestamp and image data are assigned based on minimal time difference within a threshold for each connected camera separately such as 
\begin{equation}
\begin{split}
    o_{cam} = &n_{image}(t_{image} \approx t_{VersaVIS}) -\\ &n_{VersaVIS}(t_{image} \approx t_{VersaVIS}),
    \end{split}
\end{equation}
where $o_{cam}$ is the sequence number offset.
As the images are triggered very slowly (e.g. $\unit[1]{Hz}$), the \ac{usb} buffer and \ac{os} scheduling jitter is assumed to be negligible.\\
As soon as $o_{cam}$ is constant and time offsets are small, the trigger board is notified about the initialization status of the camera. With all cameras initialized, normal triggering mode (e.g. high frequency) can be activated.\\
During normal mode, image data (directly from the camera) and image timestamps (from \ac{VersaVIS} triggering board) are associated based on sequence number like
    \begin{equation}
        t_{image} = t_{VersaVIS}(n_{image}+o_{cam}).
    \end{equation}
\subsubsection{\ac{imu} receiver}
In addition to the camera data, in a setup where the \ac{imu} is triggered and read out by the \ac{VersaVIS} triggering board, the \ac{imu} message should only hold minimal information to minimize bandwidth requirements and therefore needs to be reassembled into a full \ac{imu} message on the host computer.

\subsection{\ac{VersaVIS} triggering board}
One main part of \ac{VersaVIS} is the triggering board which is a \ac{mcu}-based custom \ac{pcb} shown in Figure~\ref{fig:versavis} that is used to connect all sensors and performs sensor synchronization. For easy extensibility and integration, the board is compatible with the Arduino environment \cite{ArduinoInc.2019ArduinoZero}. In the reference design, the board supports up to three independently triggered cameras with a four pin connector. Furthermore, \ac{spi}, \ac{i2c} or \ac{uart} can be used to interface with an \ac{imu} or other sensors. Table~\ref{tab:pcb} shows the specifications of the triggering board. The board is connected to the host computer using \ac{usb} and communicates with \ac{ros} using \textit{rosserial} \cite{Ferguson2019Rosserial}.
\begin{table}[]
    \centering
    \caption{Hardware characteristics for the \ac{VersaVIS} triggering board.}
    \label{tab:pcb}
    \begin{tabular}{l|l}
    \hline
        \ac{mcu}& ARM M0+ \\
        Hardware interface & \ac{spi}, \ac{i2c}, \ac{uart}\\
        Host interface & Serial-\ac{usb} 2.0\\
        Weight & $\unit[15.2]{g}$\\
        Size & $\unit[62\times 40\times 13.4]{mm}$ \\
        Price & $<\unit[100]{\$}$\\ \hline
    \end{tabular}
\end{table}

\section{Evaluations}\label{sec:evaluations}
In this section, several evaluations are carried out that show the synchronization accuracy of different modules of the \ac{VersaVIS} framework.
\subsection{Camera-Camera}
\begin{figure}
\includegraphics[width=\columnwidth]{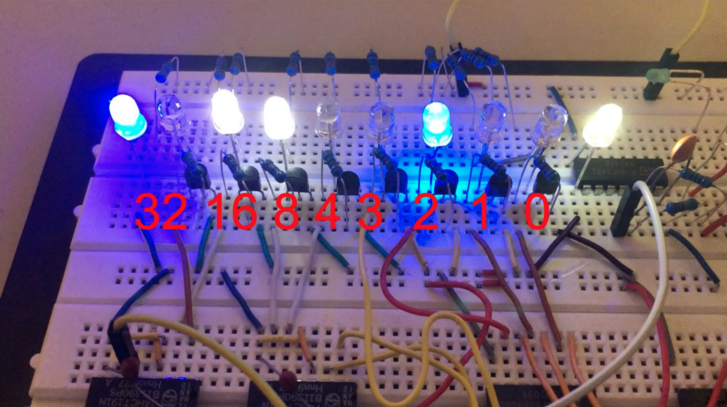}
\caption{\ac{led} timing board with indicated state numbers. The left and right most \acp{led} are used for position reference. The four right counting \acp{led} are striding while the four counting \acp{led} on the left side are binary encoded.}
\label{fig:led-board}
\end{figure}

\begin{figure}
    \centering
        \subfloat[\ac{led} count $c_1=16$] {%
            \includegraphics[width=0.8\columnwidth, clip, valign=m]{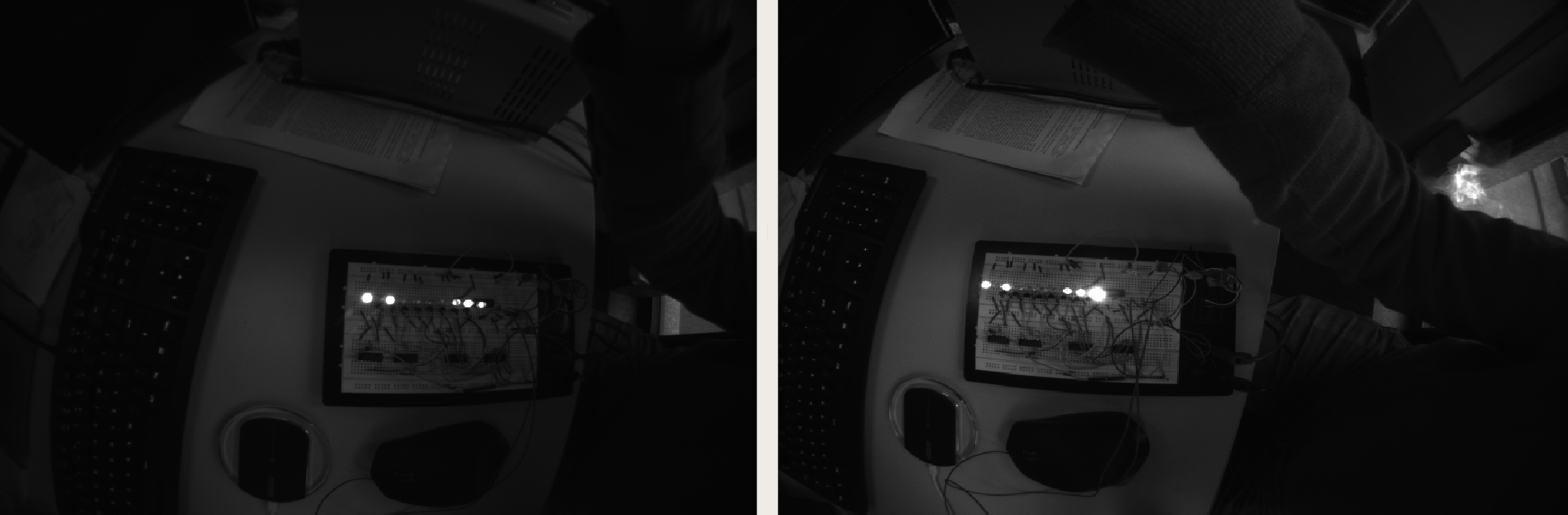}
            \label{fig:camcam1}
        }\\
        \subfloat[\ac{led} count $c_2= c_1+ k= 116 \% 64 = 52$] {%
            \includegraphics[width=0.8\columnwidth, valign=m]{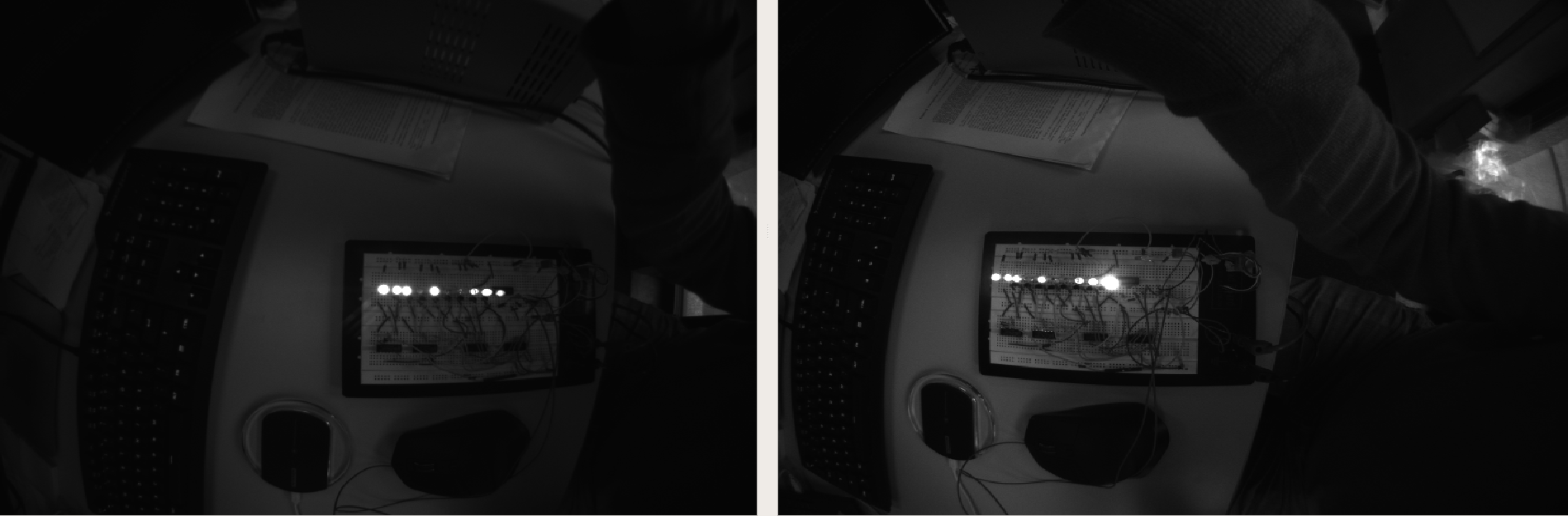}
            \label{fig:camcam2}
        } \\
        \subfloat[\ac{led} count $c_3 = c_2 + k = 216 \% 64 = 24$] {%
            \includegraphics[width=0.8\columnwidth, valign=m]{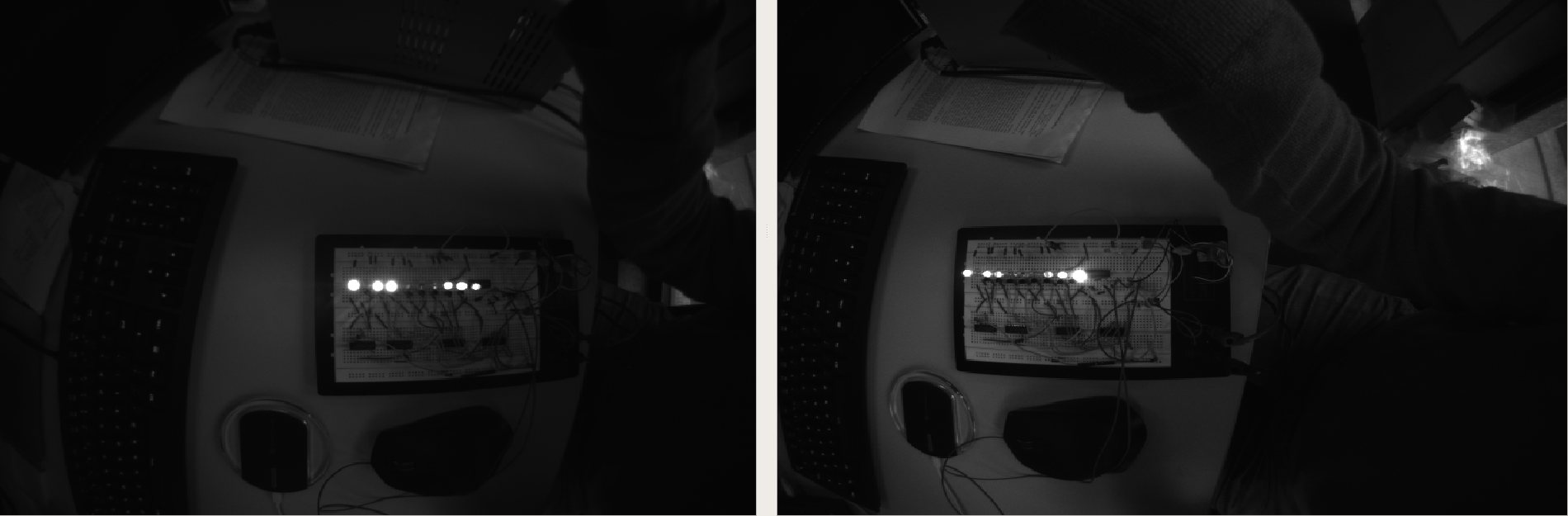}
            \label{fig:camcam3}
        }
        \caption{Measurements of two independently triggered synchronized cameras with exposure compensation (left \texttt{cam0}, right \texttt{cam2}). Both cameras show strictly the same image with at most two of the striding \acp{led} on. The increment between consecutive measurements adds up correctly and overflows at a count of $64$.}
        \label{fig:camcam}
    \end{figure}

The first important characteristic of a good multi-cam time synchronization is that multiple corresponding camera images capture the same information. This is especially important when multiple cameras are used for state estimation (see Section~\ref{sec:trains}). \\
For the purpose of evaluating the synchronization accuracy of multi-camera triggering, we captured a stream of images of an \ac{led} timing board shown in Figure~\ref{fig:led-board} with two independently triggered but synchronized and exposure-compensated cameras.
The board features eight counting \acp{led} (the left and right most \acp{led} are always on and used for position reference). The board is changing state whenever a trigger is received. The \acp{led} are organized in two groups. The right group of four indicate one count each, while the left group is binary encoded resulting in a counter overflow at $64$. The board is triggered with $f_l = \unit[1]{kHz}$ aligned with the mid-exposure timestamps of the images. Furthermore, both cameras are operated at a rate of $f_c = \unit[10]{Hz}$ and with a fixed exposure time of $t_c = \unit[1]{ms}$. \\
For a successful synchronization of the sensors, images captured with both cameras should show the same bit count $c$ with at most two of the striding \acp{led} on (since the board changes state at mid-exposure) and also the correct increment between images of $k = \frac{f_l}{f_c} = 100$. \\
Figure~\ref{fig:camcam} shows results of three consecutive image pairs. All image pairs (left and right) show the same \ac{led} count while consecutive images show the correct increment of $100$. An image stream containing  $400$ image pairs was inspected without any case of wrong increment or non-matching pairs.\\
We can therefore conclude that the time synchronization of two cameras has an accuracy better than $\frac{1}{2 \cdot f_l} = \unit[0.5]{ms}$ showing an accurate time synchronization.
\subsection{Camera-IMU}
\begin{figure*}
\centering
\includegraphics[width=1.8\columnwidth]{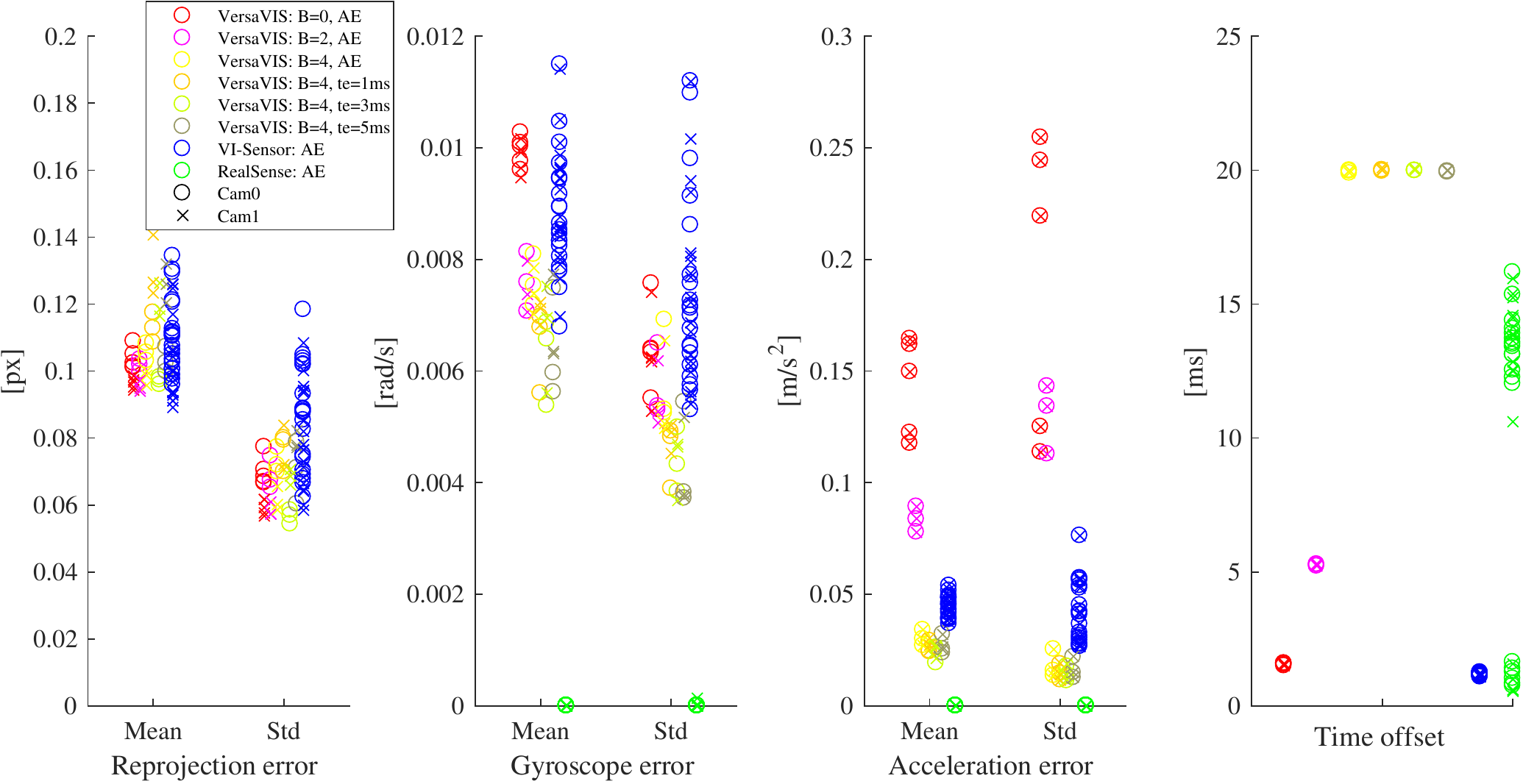}
\caption{Distribution of camera to \ac{imu} calibration results using the different sensors and different sensor configurations for \ac{VersaVIS}.}
\label{fig:camimu}
\end{figure*}

For many \ac{vio} algorithms such as ROVIO \cite{Bloesch2015} or OKVIS \cite{Leutenegger2015}, accurate time synchronization of camera image and \ac{imu} measurement is crucial for a robust and accurate operation. \\
Typically, time offsets between camera and \ac{imu} can be the result of data transfer delay, \ac{os} scheduling, clock offsets of measurement devices, changing exposure times or internal filtering of \ac{imu} measurements. Thereby, only offsets that are not constant are critical as constant offsets can be calibrated. Namely, offsets that are typically changing such as \ac{os} scheduling, clocks on different measurement devices or a not compensated changing exposure time should be avoided.\\
Using the camera-IMU calibration framework Kalibr \cite{Furgale2013UnifiedSystems}, a time offset between camera and \ac{imu} measurement can be determined by optimizing the extrinsic calibration together with a constant offset between both modalities. To test the consistency of the time offset, multiple datasets were recorded with \ac{VersaVIS} using different configurations for the \ac{imu} filtering and exposure times. The window width $B$ of the deployed Barlett window \ac{fir} filter on the \ac{imu} \cite{AnalogDevices2019ADIS16448BLMZ:Sensor} was set to $B=\{0,2,4\}$ while the exposure time $t_e$ was set to \ac{ae} or fixed to $t_e=\unit[\{1,3,5\}]{ms}$. \\
Furthermore, also the Skybotix VI-Sensor \cite{Nikolic2014} and Intel RealSense T265 \cite{IntelCorporation2019IntelT265} were tested for reference.\\
Table~\ref{tab:camimu} and Figure~\ref{fig:camimu} show import indicators of the calibration quality for multiple datasets. The reprojection error represents how well the lens and distortion model fit the actual lens and how well the movement of the camera agrees with the movement of the \ac{imu} after calibration. The reprojection error of \ac{VersaVIS} and VI-Sensor are both low and consistent meaning that the the calibration converged to a consistent extrinsic transformation between camera and \ac{imu} and both sensor measurements agree well. Furthermore, the reprojection errors are independent of the filter and exposure time configuration showing that exposure compensation is working as expected. On the other side, the RealSense shows high reprojection errors because of the sensor's fisheye lenses which turn out to be hard to calibrate even with the available fisheye lens models \cite{Usenko2018TheModel} in Kalibr.\\
This also becomes visible in the acceleration and gyroscope errors where the errors are very low as a result to the poorly fitting lens model. Kalibr estimates the body spline to be purely represented by the \ac{imu} and neglect image measurements. For \ac{VersaVIS} both errors are highly dependent on the \ac{imu} filter showing a decrease of error with more aggressive filtering due to minimized noise. However, also here similar or lower errors of \ac{VersaVIS} can be achieved compared to the VI-Sensor. \\
Finally, the time offset between camera and \ac{imu} measurements shows that \ac{VersaVIS} and VI-Sensor possess a similar accuracy in time synchronization as the standard deviations are low and the time offsets consistent indicating synchronization accuracy below $\unit[0.05]{ms}$. However, time offset is highly dependent on the \ac{imu} filter configuration. Therefore, the delay should be compensated either on the driver side or on the estimator side when more aggressive filtering is used (e.g. to reduce the influence of vibrations). Furthermore, the time offset is independent of exposure time and camera indicating again that exposure compensation is working as intended. RealSense shows inconsistent time offset estimations with a bi-modal distribution delimited by half the inter-frame time of $\approx \unit[15]{ms}$ indicating that for some datasets, there might be image measurements shifts by one frame.
\begin{table*}[t]
    \centering
        \caption{Mean values of camera to \ac{imu} calibration results for different sensors and sensor configurations for \ac{VersaVIS}.}
        \label{tab:camimu}
    \begin{tabular}{ccc|c|cc|cc|cc|cc}
        &$B$&$t_e$&N& \multicolumn{2}{c|}{Reprojection error $[\unit{px}]$}  & \multicolumn{2}{c|}{Gyroscope error $[\unitfrac{rad}{s}]$} & \multicolumn{2}{c|}{Acceleration error $[\unitfrac{m}{s^2}]$} & \multicolumn{2}{c}{Time offset $[\unit{ms}]$} \\ 
                        &&&& Mean  & Std   & Mean  & Std& Mean  & Std& Mean & Std\\\hline
VersaVIS &$0$ &\ac{ae} & 10 & 0.100078 & 0.064657 &0.009890 &0.006353 &0.143162 &0.191270 &1.552360 &0.034126 \\
VersaVIS &$2$& \ac{ae} & 6 & $\mathbf{0.098548}$ & 0.063781 &0.007532 &0.005617 &0.083576 &0.130018 &5.260927 &0.035812 \\ 
VersaVIS &$4$& \ac{ae} & 6 & 0.101866 & 0.067196 &0.007509 &0.005676 &0.030261 &0.018168 &19.951467 &0.049712 \\
VersaVIS &$4$& $\unit[1]{ms}$ & 6 & 0.121552 & 0.075939 &0.006756 &0.004681 &0.026765 &0.014890 &20.007137 &0.047525 \\
VersaVIS &$4$& $\unit[3]{ms}$ & 6 & 0.108760 & $\mathbf{0.062456}$ &0.006483 &0.004360 &0.024309 &0.014367 &20.002966 &0.035952 \\
VersaVIS &$4$& $\unit[5]{ms}$ & 6 & 0.114614 & 0.074536 &0.006578 &0.004282 &0.027468 &0.016553 &19.962924 &$\mathbf{0.029872}$ \\
VI-Sensor &$\times$&\ac{ae}& 40 & 0.106839 & 0.083605 &0.008915 &0.007425 &0.044845 &0.042446 &$\mathbf{1.173100}$ &0.046410 \\
Realsense &$\times$&\ac{ae}& 40 & 0.436630 & 0.355895 & $\mathbf{0.000000}$ & $\mathbf{0.000005}$ & $\mathbf{0.000000}$ & $\mathbf{0.000002}$ &9.884808 &5.977421 \\
    \end{tabular}
    \label{tab:my_label}
\end{table*}
\subsection{VersaVIS-Host}
\begin{figure}
    \centering
    \includegraphics[width=\columnwidth]{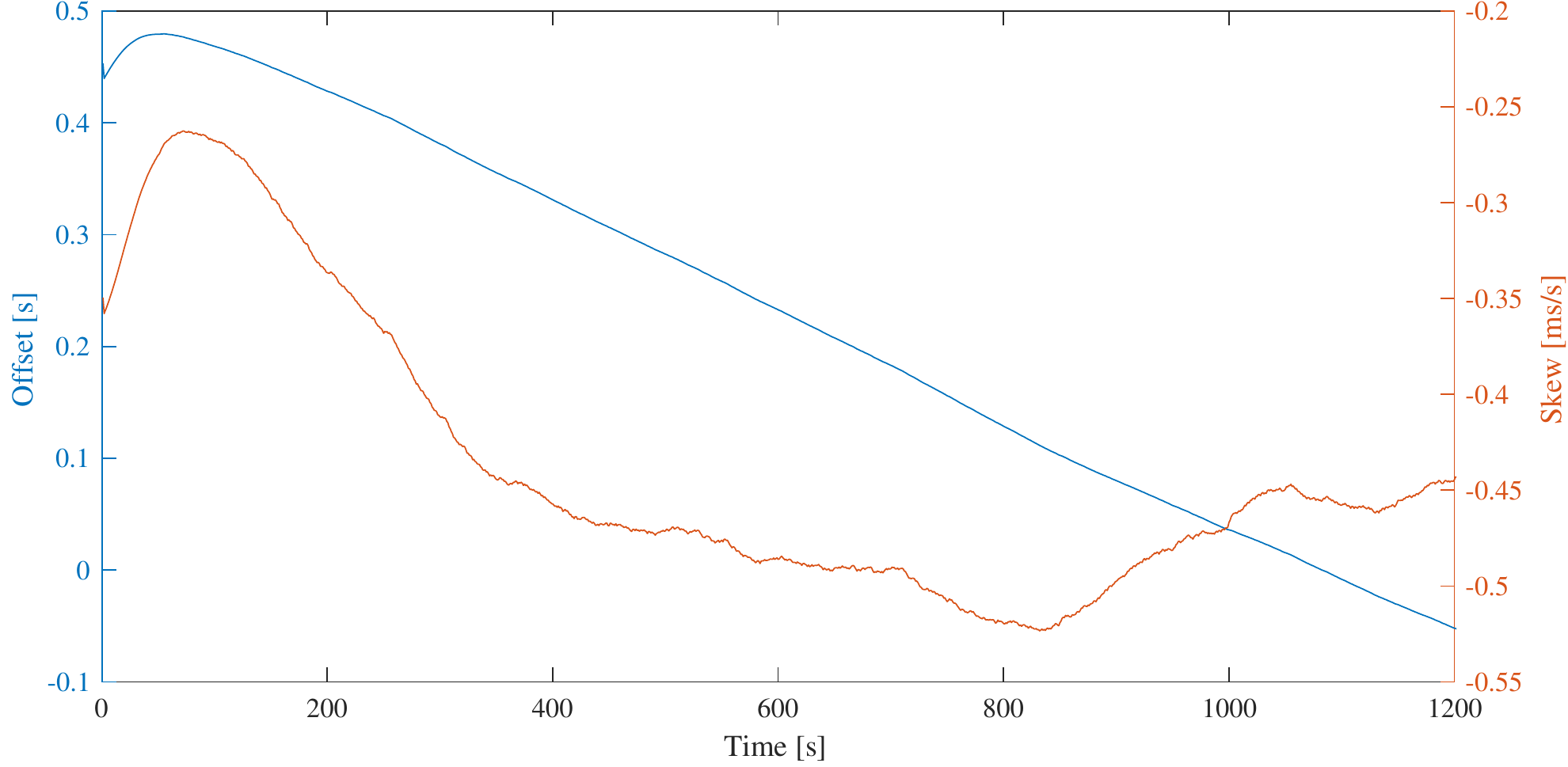}
    \caption{\ac{ekf} filter states after startup and in convergence. The offset is constantly decreasing after convergence due to a clock skew difference between \ac{VersaVIS} and host computer.}
    \label{fig:ekf_states}
\end{figure}

\begin{figure}
    \centering
    \includegraphics[width=\columnwidth]{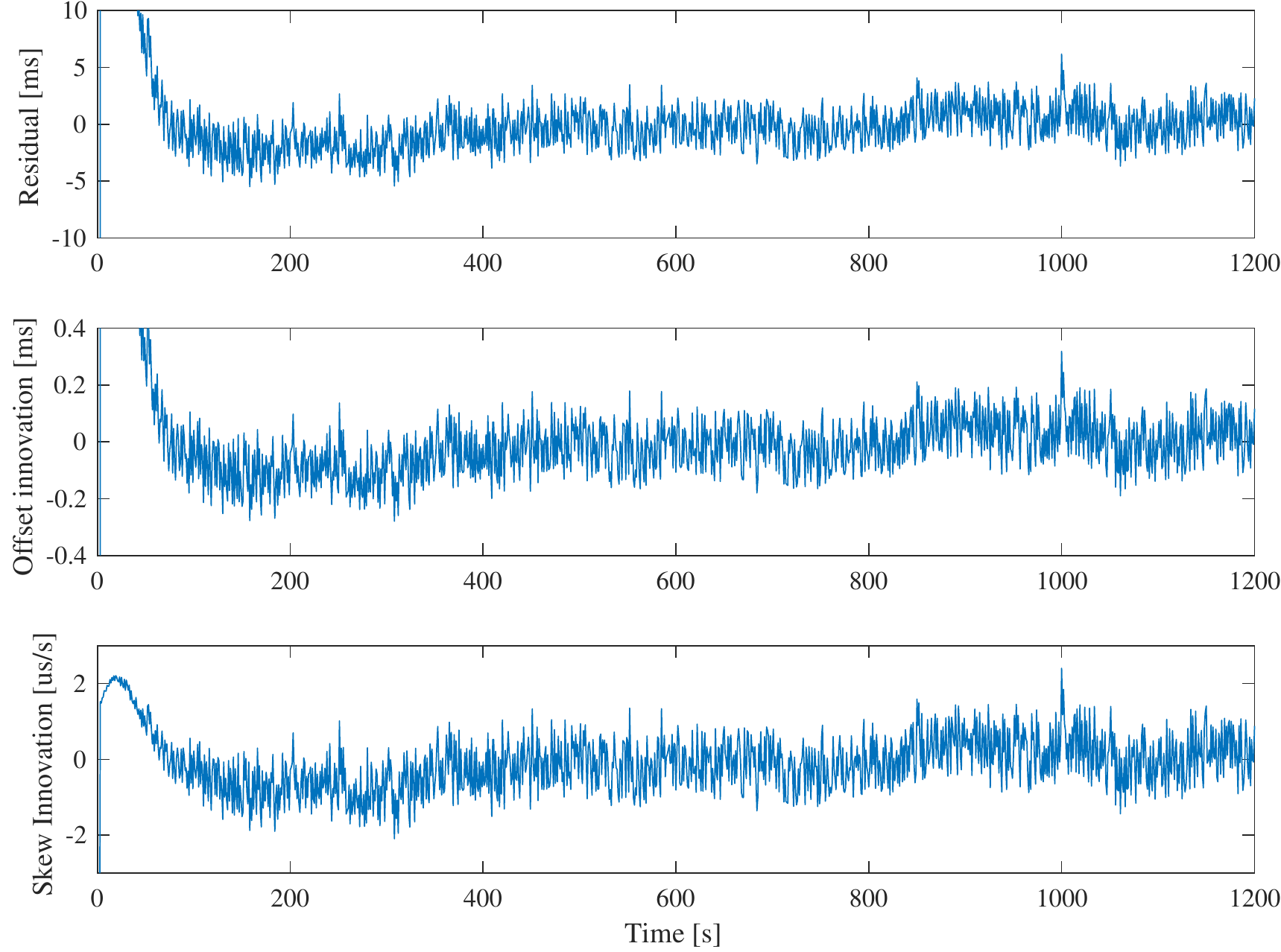}
    \caption{\ac{ekf} residual and innovation terms after startup and in convergence. The jitter in the serial-\ac{usb} interface is directly influencing the residual resulting in oscillating errors. However, as this jitter has zero-mean, the achieved clock synchronization/translation has a much higher accuracy visible in the innovation term of the clock offset.}
    \label{fig:ekf_zoomed}
\end{figure}

As mentioned in Section~\ref{sec:add_sensors}, not all sensors are directly compatible with \ac{VersaVIS}. The better the clock translation of different measurement devices, the better the sensor fusion.\\
Thanks to the bi-directional connection between \ac{VersaVIS} and host computer, clock translation requests can be sent from \ac{VersaVIS} and the response from the host can be analyzed. \\
Such requests are sent every second. Figure~\ref{fig:ekf_states} shows the evolution of the \ac{ekf} states introduced in Section~\ref{sec:add_sensors}. In this experiment, there is a clock skew between the host computer and \ac{VersaVIS} resulting in a constantly decreasing offset after \ac{ekf} convergence. This highlights the importance of estimating the skew when using time translation. \\
Figure~\ref{fig:ekf_zoomed} shows results of the residual $\epsilon$ and the innovation terms of the clock offset $\delta$ and skew $\eta$ after startup and in convergence. After approximately $\unit[60]{s}$, the residual drops below $\unit[5]{ms}$ and keeps oscillating at $\pm \unit[5]{ms}$ due to \ac{usb} jitter. However, thanks to the \ac{ekf}, this error is smoothed to a zero mean innovation of the offset of $\pm \unit[0.2]{ms}$ resulting in a clock translation accuracy of $\pm \unit[0.2]{ms}$. The influence of the skew innovation can be neglected with the short update time of $\unit[1]{s}$.
\section{Applications}\label{sec:applications}
This section validates the flexibility, robustness and accuracy of our system with several different sensor setups using \ac{VersaVIS}, utilized in different applications.
\subsection{Visual-inertial SLAM}
\label{sec:vio}
\begin{figure}
    \centering
    \includegraphics[width=\columnwidth]{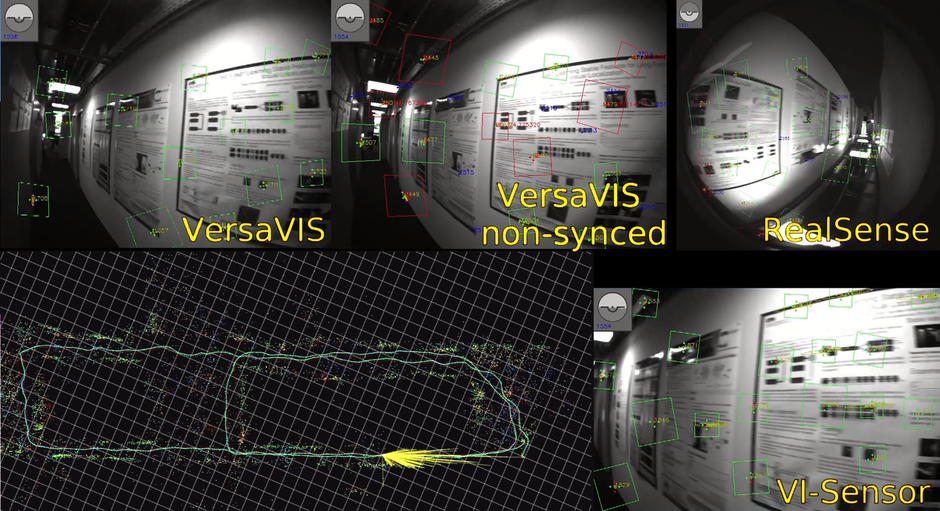}
    \caption{Feature tracking of ROVIO \cite{Bloesch2015} on a dataset recorded in our lab. Inliers of the feature tracking are shown in green while outliers are shown in red.}
    \label{fig:vio-tests}
\end{figure}

\begin{figure}
    \centering
    \includegraphics[width=\columnwidth]{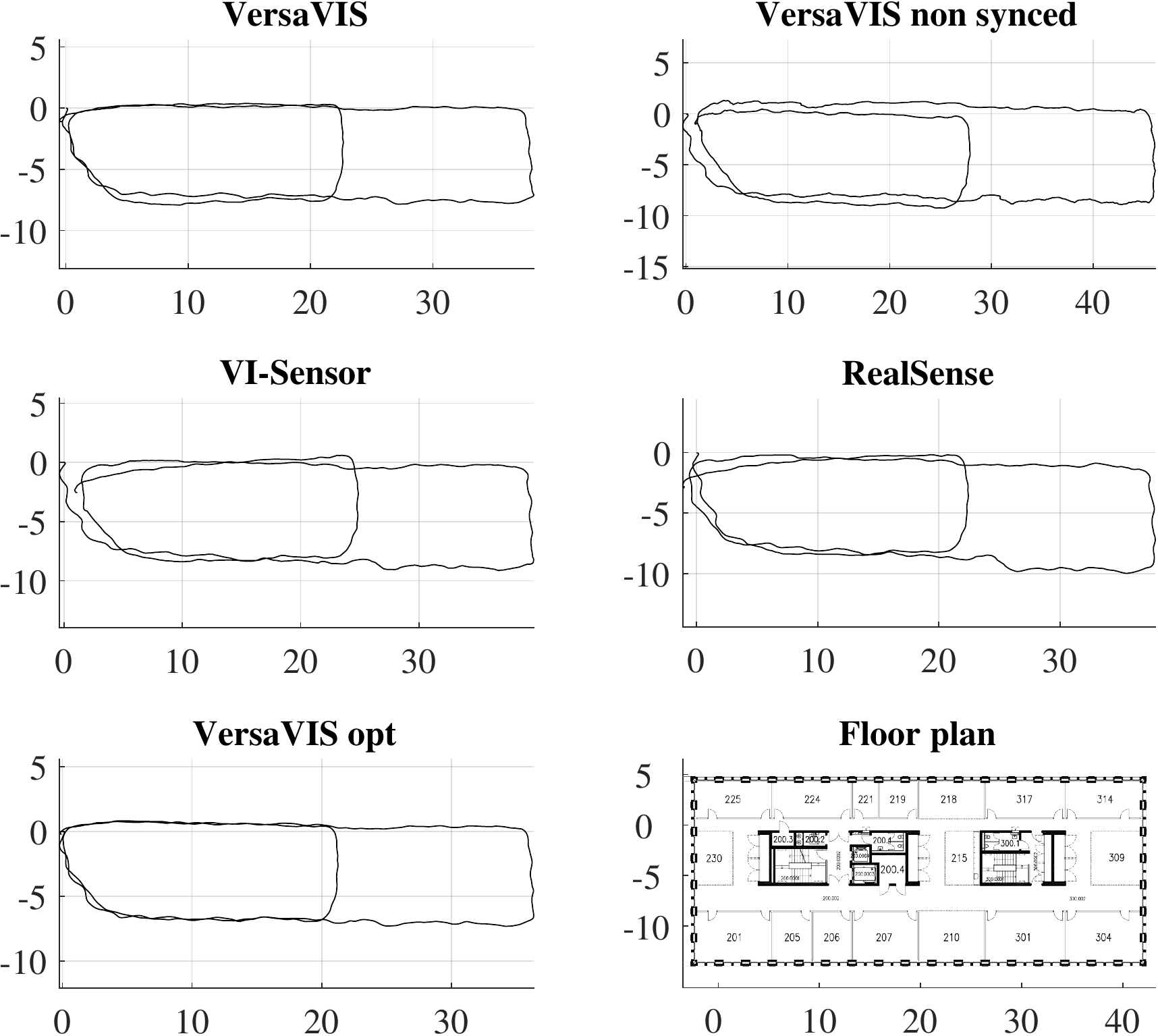}
    \caption{Trajectories of different sensor setups using ROVIO \cite{Bloesch2015} on a dataset recorded in our lab. While all of the sensors provide a useful output, \ac{VersaVIS} shows the lowest amount of drift while the non-synchronized sensor shows a lot of jitter in the estimation.}
    \label{fig:vio-tests-trajectory}
\end{figure}

The main purpose of a \ac{vi} sensor is to perform odometry estimation and mapping. For that purpose, we collected a dataset walking around in our lab with different sensor setups including \ac{VersaVIS} equipped with a FLIR BFS-U3-04S2M-CS camera and the Analog Devices ADIS16448 \ac{imu} shown in Figure~\ref{fig:lidarstick}, a Skybotix VI-Sensor \cite{Nikolic2014} and an Intel RealSense T265 all attached to the same rigid body. For reference, we also evaluated the use of the FLIR camera together with the \ac{imu} of the VI-Sensor as a non-synchronized\footnote{Since both sensors do \textit{some} time translation to the host, software time translation is available.} sensor setup.\\
Figure~\ref{fig:vio-tests} shows an example of the feature tracking window of ROVIO on the dataset. Both \ac{VersaVIS} and the VI-Sensor show many well tracked features even during fast motions depicted on the image. Due to the high \ac{fov} of the RealSense cameras and not perfectly fitting lens model, some of the keypoints do not reproject correctly to the image plane resulting in falsely warped keypoints. With a non-synchronized sensor (VersaVIS non-synced), many of the keypoints cannot be correctly tracked as the \ac{imu} measurements and the image measurements do not agree well.\\
Figure~\ref{fig:vio-tests-trajectory} shows trajectories obtained with the procedure described above. While all sensors provide useful output depending on the application, \ac{VersaVIS} shows the lowest drift. VI-Sensor and RealSense suffer from their specific camera hardware where the VI-Sensor has inferior lenses and camera chips (visible in motion blur) and RealSense has camera lenses where no well-fitting lens model is available in the used frameworks. Without synchronization, the trajectory becomes more jittery resulting in potentially unstable estimator also visible in the large scale offset and shows higher drift. Using batch optimization and loop closure \cite{Schneider2017}, the trajectory of \ac{VersaVIS} can be further optimized (VersaVIS opt). However, the discrepancy between \ac{VersaVIS} and \ac{VersaVIS} opt is small indicating an already good odometry performance.
\subsection{Stereo Visual-Inertial Odometry on Rail vehicle} \label{sec:trains}
The need for public transportation is heavily increasing while current infrastructure is reaching its limits. Using on-board sensors with reliable and accurate positioning, the efficiency of the infrastructure could be highly increased \cite{Tschopp2019ExperimentalVehiclesb}. However, this requires the fusion of multiple independent positioning modalities of which one could be visual-aided odometry.\\
For this purpose, \ac{VersaVIS} was combined with two global-shutter cameras arranged in a fronto-parallel stereo setup using master-slave triggering (see Section~\ref{sec:firmware}) and a compact, precision six degrees of freedom \ac{imu} shown in Figure~\ref{fig:stereo}. 
In comparison to many commercial sensors, the camera has to provide a high frame-rate to be able to get a reasonable number of frames per displacement, even at higher speeds and feature a high dynamic range to deal with the challenging lighting conditions. Furthermore, due to the constraint motion of the vehicle and low \ac{snr}, the \ac{imu} should be of a high quality and also should be temperature calibrated to deal with temperature changes due to direct sunlight.
The sensor specifications are summarized in Table~\ref{tab:vi-sensor}. \\
\begin{table}
\caption{Sensor specifications deployed for data collection on rail vehicles \cite{Tschopp2019ExperimentalVehiclesb}}\label{tab:vi-sensor}
\begin{tabular}{m{1cm}|m{1.8cm}|m{4.8cm}} 
Device&Type&Specification \\ \hline \hline
Camera& Basler acA1920-155uc&\makecell[l]{Frame-rate $\unit[20]{fps}$\footnotemark[4], \\Resolution $1920 \times 1200$, \\Dynamic range $\unit[73]{dB}$}\\ \hline
Lense& Edmund Optics&\makecell[l]{Focal length $\unit[8]{mm} \approx 70 \deg$ opening \\ angle; Aperture $f/5.6$}\\ \hline
IMU& ADIS16445& \makecell[l]{Temperature calibrated, $\unit[300]{Hz}$, \\ $\unitfrac[\pm250]{\deg}{s}$, $\unitfrac[\pm 49]{m}{s^2}$}\\ 
\end{tabular}
\end{table}
\footnotetext[4]{The hardware is able to capture up to $\unit[155]{fps}$.}
The results obtained with VersaVIS  (details found in our previous work \cite{Tschopp2019ExperimentalVehiclesb}) show that by using stereo cameras and tightly synchronized \ac{imu} measurements we can improve robustness and provide accurate odometry up to $\unit[1.11]{\%}$ of the travelled distance on railway scenarios and speeds up to $\unitfrac[52.4]{km}{h}$.
\subsection{Multi-modal mapping and reconstruction}
\ac{vi} mapping as described in~\ref{sec:vio} can provide reliable pose estimates in many different environments.
However, for applications that require precise mapping of structures in GPS denied, visually degraded environments, such as mines and caves, additional sensors are required.
The multi-modal sensor setup as seen in Figure~\ref{fig:lidarstick} was specifically developed for mapping research in these challenging conditions.
The fact that \ac{VersaVIS} provides time synchronization to the host computer greatly facilitates the addition of other sensors.
The prerequisite is that these additional sensors are time synchronized with the host computer as well, which in our case, an Ouster OS-1 64-beam \ac{lidar}, is done over \ac{ptp}.
The absence of light in these underground environments also required the addition of an artificial lighting source that fulfills very specific requirements to support the camera system for pose estimation and mapping.
The main challenge was to achieve the maximum amount of light, equally distributed across the environment (i.e. \textit{ambient light}), while at the same time being bound by power and cooling limitations.
To that end a pair of high-powered, camera-shutter-synchronized \acp{led}, similar to to the system shown by Nikolic~et.~al.~\cite{Nikolic2013AFacilities}, are employed.
\ac{VersaVIS} provides a trigger signal to the \ac{led} control board, which represents the union of all camera exposure signals, ensuring that all images are fully illuminated while at the same time minimizing the power consumption and heat generation.
This allows operating the \acp{led} at a significantly higher brightness level than during continuous operation.

\begin{figure}
\centering
    \subfloat[Accumulated points clouds using only \ac{vi} \ac{slam} poses.] {%
    \centering
        \includegraphics[width=0.97\columnwidth, trim={0cm 3.5cm 0cm 0cm}, clip, valign=m]{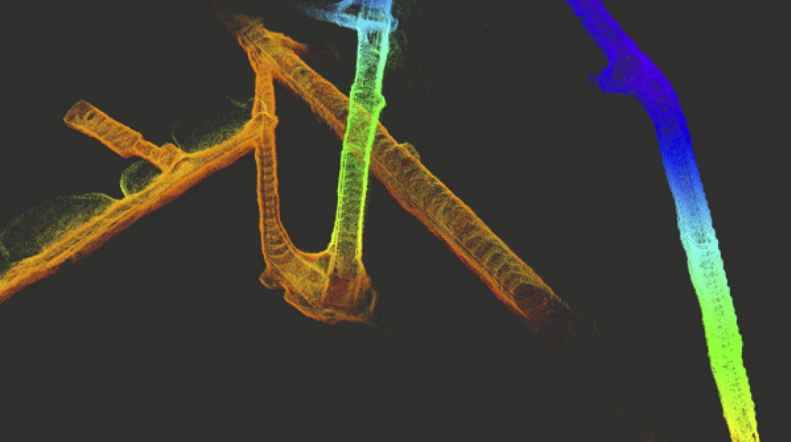}
        \label{fig:gonzen_accumulated}
    }\\
    \subfloat[Dense reconstruction example based on \cite{Oleynikova2017Voxblox:Planning}.] {%
    \centering
        \includegraphics[width=0.46\columnwidth, valign=m]{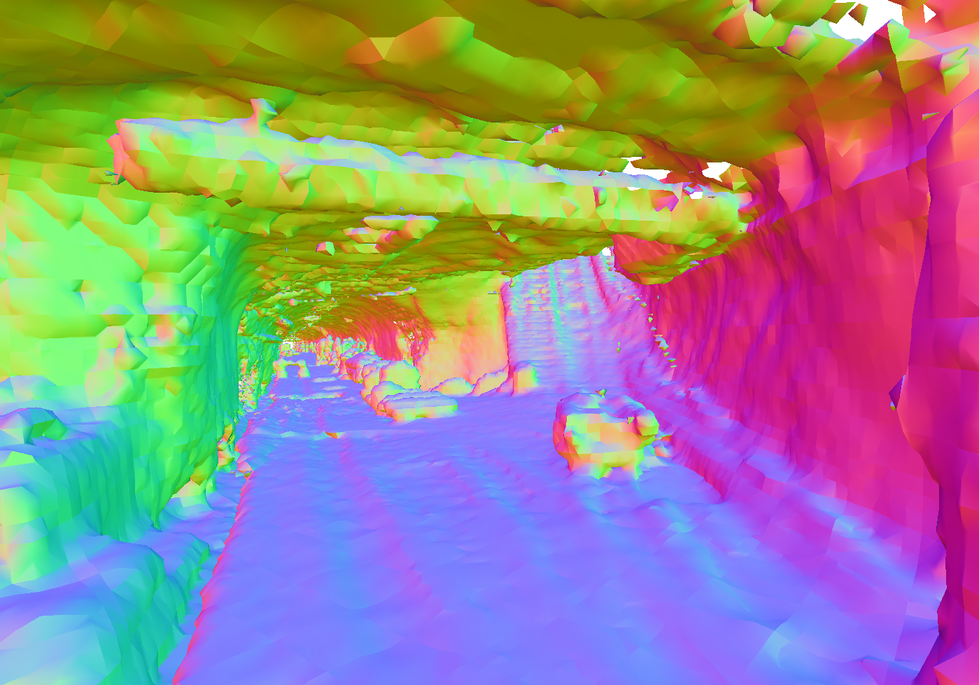}
        \label{fig:dense_1}
    } \;
    \subfloat[Visible details of wheels lying on the ground using dense reconstruction based on \cite{Oleynikova2017Voxblox:Planning}.] {%
    \centering
        \includegraphics[width=0.46\columnwidth, valign=m]{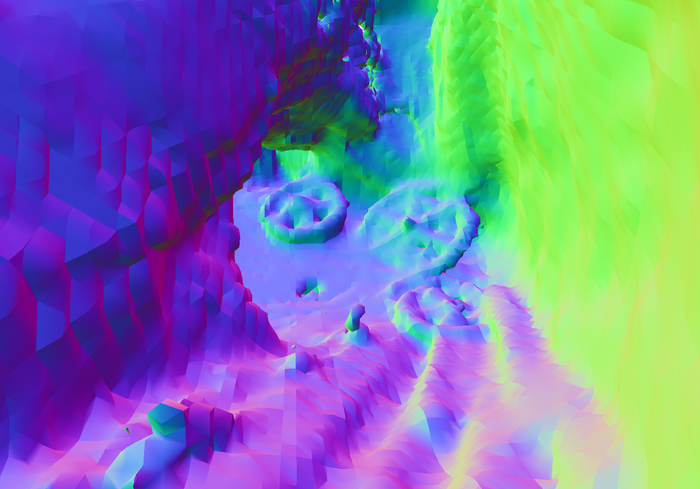}
        \vphantom{\includegraphics[width=0.46\columnwidth, valign=m]{figures/pipe200_L02.png}}%
        \label{fig:dense_2}
    }
    \caption{3D reconstruction of an underground mine in Gonzen CH based on \ac{lidar} data with poses provided by \ac{vi} \ac{slam} \cite{Schneider2017}.}
    \label{fig:gonzen}
\end{figure}

Figure~\ref{fig:gonzen} shows an example of the processed multi-modal sensor data.
\ac{vi} odometry~\cite{Bloesch2015} and mapping~\cite{Schneider2017} including local refinements based on \ac{lidar} data and \ac{tsdf}-based surface reconstruction~\cite{Oleynikova2017Voxblox:Planning} were used to precisely map the 3D structure of parts of an abandoned iron mine in Switzerland.

\subsection{Object Based Mapping}
\begin{figure}
\centering
    \subfloat[Scene reconstruction using the RGB-D-I sensor.] {%
    \centering
        \includegraphics[width=0.97\columnwidth, trim={3cm 0cm 2cm 2cm}, clip, valign=m]{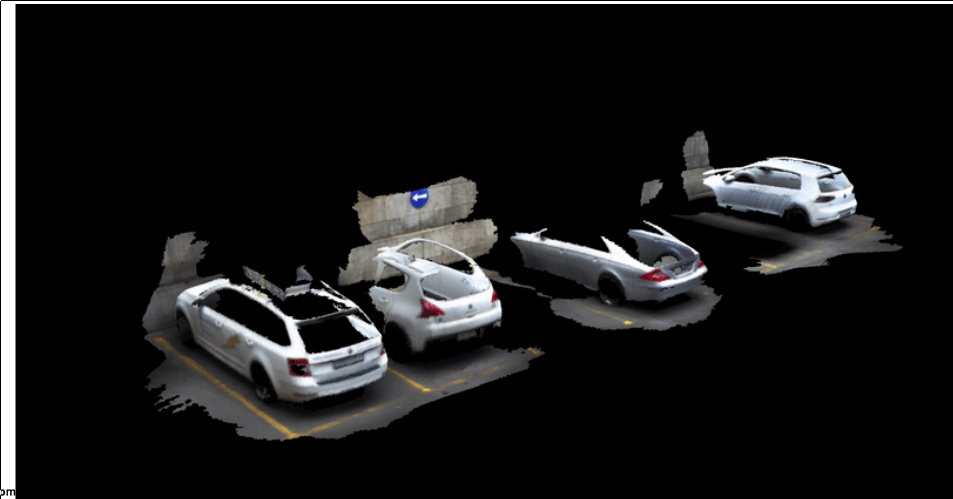}
        \label{fig:modelfiy_reconstruction}
    }\\
    \subfloat[Segmented objects in a parking garage using \cite{Furrer2019Modelify:Completion}.] {%
    \centering
        \includegraphics[width=0.465\columnwidth, valign=m]{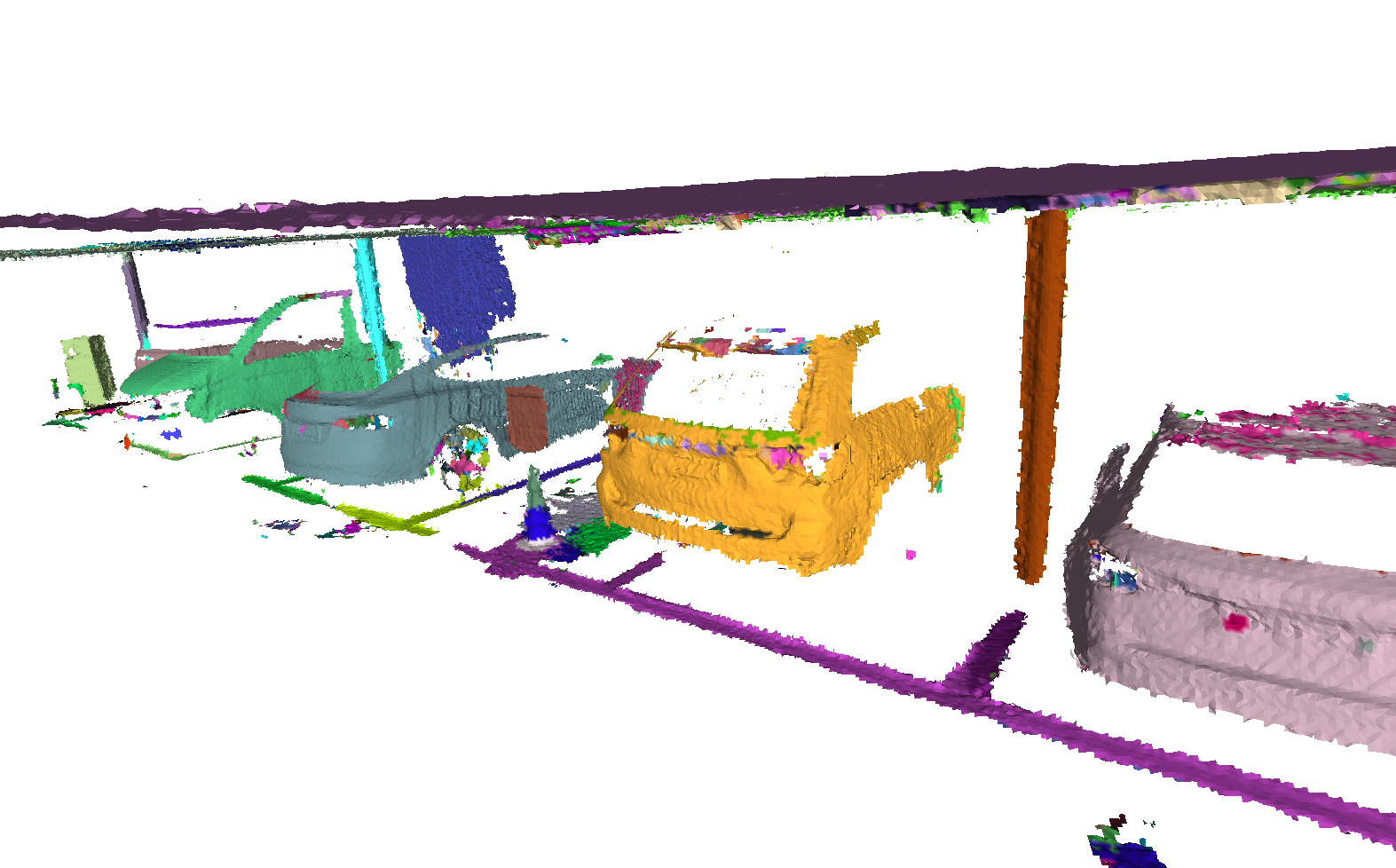}
         \vphantom{\includegraphics[width=0.46\columnwidth, valign=m]{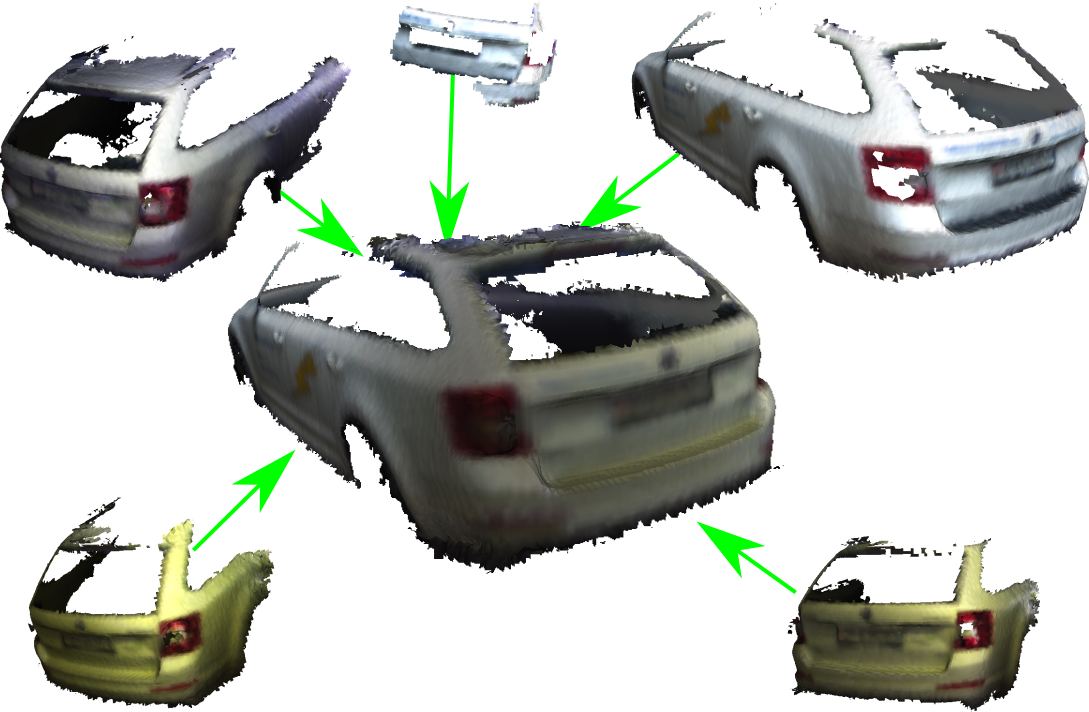}}%
        \label{fig:gsm_garage_103}
    } \;
    \subfloat[Merging of objects from the database \cite{Furrer2019Modelify:Completion}.] {%
    \centering
        \includegraphics[width=0.465\columnwidth, valign=m]{figures/modelify_merging.png}
        \label{fig:merged_objects}
    }
    \caption{Object segmentation and reconstruction based on \cite{Furrer2019Modelify:Completion} and \cite{Furrer2018IncrementalObservations} using data from the RGB-D-I sensor and camera poses obtained with maplab \cite{Schneider2017}.}
    \label{fig:modelify}
\end{figure}

Robots that operate in changing environments benefit from using maps that are based on physical objects instead of abstract points.
Such object based maps are more consistent if individual objects move, as only a movement of an object must be detected instead of each point on the moved object that is part of the map.
Object based maps are also a better representation for manipulation tasks.
The elements in the map are typically directly the objects of interest in such tasks.\\
For the object based mapping application, a sensor setup with a depth and an RGB camera, and an \ac{imu} was assembled, see Figure~\ref{fig:rgbdi}.
A Pico Monstar, which is a \ac{tof} camera that provides $352 \times 287$ resolution depth images at up to $\unit[60]{Hz}$, was combined with a Flir Blackfly $\unit[1.6]{MP}$ color camera, and an ADIS16448 \ac{imu}.
To obtain accurate pose estimates, the \ac{imu} and color camera were used for odometry and localization~\cite{Schneider2017}.\\
With these poses, and together with the depth measurements, an approach from~\cite{Furrer2019Modelify:Completion} was used to reconstruct the scene and extract object instances.
An example of such a segmentation map is shown in Figure~\ref{fig:modelfiy_reconstruction}, and objects that were extracted and inserted into a database in Figure~\ref{fig:gsm_garage_103} and \ref{fig:merged_objects}.

\section{Conclusions}\label{sec:conclusions}
We presented a hardware synchronization suite for multi-camera \ac{vi} sensors consisting of the full hardware design, firmware and host driver software. The sensor suite supports multiple beneficial features such as exposure time compensation and host time translation and can be used in both independent and master-slave multi-camera mode.\\ 
The time synchronization performance is analyzed separately for camera-camera synchronization, camera-\ac{imu} synchronization and \ac{VersaVIS}-host clock translation. All modules achieve time synchronization accuracy of $<\unit[1]{ms}$ which is expected to be accurate enough for most mobile robotic applications.\\ 
The benefits and great versatility range of the sensor suite are demonstrated on multiple applications including hand-held \ac{vio}, multi-camera \ac{vi} applications on rail vehicles
as well as large scale environment reconstruction and object based mapping.\\
For the benefit of the community, all hardware and software components are completely open-source with a permissive license and based on easily available hardware and development software. 
This paper and the accompanying framework can also serve as a freely available reference design for research and industry as it summarizes solution approaches to multiple challenges of developing a synchronized multi-modal sensor setup. \\
The research community can easily adopt, adapt and extend this sensor setup and rapid-prototype custom sensor setups for variety of robotic applications. 
Many experimental features showcase the easy extensibility of the framework:
\begin{itemize}
    \item Illumination module: The \ac{VersaVIS} triggering board can be paired with \acp{led} shown in Figure~\ref{fig:lidarstick} which are triggered corresponding to the longest exposure time of the connected cameras. Thanks to the synchronization, the \acp{led} can be operated at a higher brightness as which would be possible in continuous operation.
    \item \ac{imu} output: The \ac{spi} output of the board enables to use the same \ac{imu} which is used in the \ac{vi} setup for a low-level controller such as the PixHawk \cite{Meier2019Pixhawk4} used in \ac{mav} control.
    \item \ac{pps} sync: Some sensors such as specific \acp{lidar} allow synchronization to a \ac{pps} signal provided by e.g. a \ac{gps} receiver or real-time clock. Using the external clock input on the triggering board, \ac{VersaVIS} can be extended to synchronize to the \ac{pps} source.
    \item \ac{lidar} synchronization: The available auxiliary interface on \ac{VersaVIS} could be used to tightly integrate \ac{lidar} measurements by getting digital pulses from the \ac{lidar} corresponding to taken measurements. The merging procedure would then be similar to the one described in Section~\ref{sec:synchronizer} for cameras with fixed exposure time.
\end{itemize}

\section*{Acknowledgement}
This work was partly supported by Siemens Mobility, Germany and by the National Center of Competence in Research (NCCR) Robotics through the Swiss National Science Foundation. The authors would like to thank Gabriel Waibel for his work adapting the firmware for a newer \ac{mcu}.

\bibliographystyle{IEEEtran}
\bibliography{references.bib}

\begin{thebibliography}{10}
\providecommand{\url}[1]{#1}
\csname url@samestyle\endcsname
\providecommand{\newblock}{\relax}
\providecommand{\bibinfo}[2]{#2}
\providecommand{\BIBentrySTDinterwordspacing}{\spaceskip=0pt\relax}
\providecommand{\BIBentryALTinterwordstretchfactor}{4}
\providecommand{\BIBentryALTinterwordspacing}{\spaceskip=\fontdimen2\font plus
\BIBentryALTinterwordstretchfactor\fontdimen3\font minus
  \fontdimen4\font\relax}
\providecommand{\BIBforeignlanguage}[2]{{%
\expandafter\ifx\csname l@#1\endcsname\relax
\typeout{** WARNING: IEEEtran.bst: No hyphenation pattern has been}%
\typeout{** loaded for the language `#1'. Using the pattern for}%
\typeout{** the default language instead.}%
\else
\language=\csname l@#1\endcsname
\fi
#2}}
\providecommand{\BIBdecl}{\relax}
\BIBdecl

\bibitem{RolandSiegwart2011}
{Roland Siegwart}, {Illah R. Nourbakhsh}, and {Davide Scaramuzza},
  \emph{{Introduction to Autonomous Mobile Robots}}, 2nd~ed., {Ronald C.
  Arkin}, Ed.\hskip 1em plus 0.5em minus 0.4em\relax Camebridge: MIT Press,
  2011.

\bibitem{Bloesch2015}
M.~Bloesch, S.~Omari, M.~Hutter, and R.~Siegwart, ``{Robust visual inertial
  odometry using a direct EKF-based approach},'' in \emph{IEEE International
  Conference on Intelligent Robots and Systems}, Seattle, WA, USA, 2015, pp.
  298--304.

\bibitem{Leutenegger2015}
S.~Leutenegger, S.~Lynen, M.~Bosse, R.~Siegwart, and P.~Furgale,
  ``{Keyframe-based visual–inertial odometry using nonlinear optimization},''
  \emph{The International Journal of Robotics Research}, vol.~34, no.~3, pp.
  314--334, 2015.

\bibitem{Qin2018VINS-Mono:Estimator}
T.~Qin, P.~Li, and S.~Shen, ``{VINS-Mono: A Robust and Versatile Monocular
  Visual-Inertial State Estimator},'' \emph{IEEE Transactions on Robotics},
  vol.~34, no.~4, pp. 1004--1020, 2018.

\bibitem{Tschopp2019ExperimentalVehiclesb}
F.~Tschopp, T.~Schneider, A.~W. Palmer, N.~Nourani-Vatani, C.~Cadena~Lerma,
  R.~Siegwart, and J.~Nieto, ``{Experimental Comparison of Visual-Aided
  Odometry Methods for Rail Vehicles},'' \emph{IEEE Robotics and Automation
  Letters}, 2019.

\bibitem{Nikolic2014}
J.~Nikolic, J.~Rehder, M.~Burri, P.~Gohl, S.~Leutenegger, P.~T. Furgale, and
  R.~Siegwart, ``{A synchronized visual-inertial sensor system with FPGA
  pre-processing for accurate real-time SLAM},'' in \emph{IEEE International
  Conference on Robotics and Automation}, Hong Kong, China, 2014, pp. 431--437.

\bibitem{IntelCorporation2019IntelT265}
\BIBentryALTinterwordspacing
{Intel Corporation}, ``{Intel{\textregistered} RealSense™ Tracking Camera
  T265},'' 2019. [Online]. Available:
  \url{https://www.intelrealsense.com/tracking-camera-t265/}
\BIBentrySTDinterwordspacing

\bibitem{Honegger2017EmbeddedStereo}
D.~Honegger, T.~Sattler, and M.~Pollefeys, ``{Embedded Real-Time Multi-Baseline
  Stereo},'' in \emph{IEEE International Conference on Robotics and
  Automation}, vol. 688007, no. 688007, Singapore, 2017.

\bibitem{ZheZhang2018}
{Zhe Zhang}, {Shaoshan Liu}, {Grace Tsai}, {Hongbing Hu}, {Chen-Chi Chu}, and
  {Feng Zheng}, ``{PIRVS: An Advanced Visual-Inertial SLAM System with Flexible
  Sensor Fusion and Hardware Co-Design},'' in \emph{ICRA}, Brisbane, 2018.

\bibitem{ArduinoInc.2019ArduinoZero}
\BIBentryALTinterwordspacing
{Arduino Inc.}, ``{Arduino Zero},'' 2019. [Online]. Available:
  \url{https://store.arduino.cc/arduino-zero}
\BIBentrySTDinterwordspacing

\bibitem{StanfordArtificialIntelligenceLaboratoryetal.2018RoboticSystem}
\BIBentryALTinterwordspacing
{Stanford Artificial Intelligence Laboratory et al.}, ``{Robotic Operating
  System},'' 2018. [Online]. Available: \url{https://www.ros.org}
\BIBentrySTDinterwordspacing

\bibitem{Furgale2013UnifiedSystems}
P.~Furgale, J.~Rehder, and R.~Siegwart, ``{Unified temporal and spatial
  calibration for multi-sensor systems},'' in \emph{IEEE International
  Conference on Intelligent Robots and Systems}, Tokyo, Japan, 2013, pp.
  1280--1286.

\bibitem{Zhang2015Visual-lidarFast}
J.~Zhang and S.~Singh, ``{Visual-lidar Odometry and Mapping: Low-drift, Robust,
  and Fast},'' in \emph{2015 IEEE International Conference on Robotics and
  Automation (ICRA)}, Seattle, Washington, 2015, pp. 2174--2181.

\bibitem{Sommer2017ACameras}
H.~Sommer, R.~Khanna, I.~Gilitschenski, Z.~Taylor, R.~Siegwart, and J.~Nieto,
  ``{A low-cost system for high-rate, high-accuracy temporal calibration for
  LIDARs and cameras},'' in \emph{IEEE International Conference on Intelligent
  Robots and Systems}, 2017.

\bibitem{Ferguson2019Rosserial}
M.~Ferguson, P.~Bouchier, and M.~Purvis, ``{rosserial},'' 2019.

\bibitem{AnalogDevices2019ADIS16448BLMZ:Sensor}
\BIBentryALTinterwordspacing
{Analog Devices}, ``{ADIS16448BLMZ: Compact, Precision Ten Degrees of Freedom
  Inertial Sensor},'' Analog Devices, Tech. Rep., 2019. [Online]. Available:
  \url{https://www.analog.com/media/en/technical-documentation/data-sheets/ADIS16448.pdf}
\BIBentrySTDinterwordspacing

\bibitem{Usenko2018TheModel}
V.~Usenko, N.~Demmel, and D.~Cremers, ``{The double sphere camera model},'' in
  \emph{Proceedings - 2018 International Conference on 3D Vision, 3DV
  2018}.\hskip 1em plus 0.5em minus 0.4em\relax Institute of Electrical and
  Electronics Engineers Inc., 10 2018, pp. 552--560.

\bibitem{Schneider2017}
T.~Schneider, M.~Dymczyk, M.~Fehr, K.~Egger, S.~Lynen, I.~Gilitschenski, and
  R.~Siegwart, ``{maplab: An Open Framework for Research in Visual-inertial
  Mapping and Localization},'' \emph{IEEE Robotics and Automation Letters},
  vol.~3, no.~3, pp. 1418--1425, 11 2018.

\bibitem{Nikolic2013AFacilities}
J.~Nikolic, M.~Burri, J.~Rehder, S.~Leutenegger, C.~Huerzeler, and R.~Siegwart,
  ``{A UAV system for inspection of industrial facilities},'' in \emph{IEEE
  Aerospace Conference}, Big Sky, MT, 2013.

\bibitem{Oleynikova2017Voxblox:Planning}
H.~Oleynikova, Z.~Taylor, M.~Fehr, R.~Siegwart, and J.~Nieto, ``{Voxblox:
  Incremental 3D Euclidean Signed Distance Fields for on-board MAV planning},''
  \emph{IEEE International Conference on Intelligent Robots and Systems}, vol.
  2017-Septe, pp. 1366--1373, 2017.

\bibitem{Furrer2019Modelify:Completion}
F.~Furrer, T.~Novkovic, M.~Fehr, M.~Grinvald, C.~Cadena, J.~Nieto, and
  R.~Siegwart, ``{Modelify: An Approach to Incrementally Build 3D Object Models
  for Map Completion},'' \emph{Manuscript submitted for publication}, 2019.

\bibitem{Furrer2018IncrementalObservations}
F.~Furrer, T.~Novkovic, M.~Fehr, A.~Gawel, M.~Grinvald, T.~Sattler,
  R.~Siegwart, and J.~Nieto, ``{Incremental Object Database: Building 3D Models
  from Multiple Partial Observations},'' in \emph{2018 IEEE/RSJ International
  Conference on Intelligent Robots and Systems (IROS)}, Madrid, Spain, 2018.

\bibitem{Meier2019Pixhawk4}
\BIBentryALTinterwordspacing
L.~Meier and {Auterion}, ``{Pixhawk 4},'' 2019. [Online]. Available:
  \url{https://docs.px4.io/v1.9.0/en/flight_controller/pixhawk4.html}
\BIBentrySTDinterwordspacing

\end{thebibliography}

\begin{acronym}
\acro{imu}[IMU]{inertial measurement unit}
\acro{etcs}[ETCS]{European Train Control System}

\acro{gps}[GPS]{Global Position System}
\acro{gnss}[GNSS]{global navigation satellite system}
\acro{asl}[ASL]{Autonomous Systems Lab}
\acro{rtk}[RTK]{real time kinematics}
\acro{slam}[SLAM]{Simultaneous Localization and Mapping}
\acro{etsc}[ETSC]{European Train Security Council}
\acro{dvs}[DVS]{Dynamic Vision Sensor}
\acro{zvv}[ZVV]{Z\"urich Verkehrsverein}
\acro{ekf}[EKF]{extended Kalman filter}
\acro{vi}[VI]{visual-inertial}
\acro{vio}[VIO]{visual-inertial odometry}
\acro{vo}[VO]{visual odometry}
\acro{lssvm}[LSSVM]{least squares support vector machine}
\acro{pf}[PF]{particle filter}
\acro{dso}[DSO]{direct sparse odometry}
\acro{mcu}[MCU]{micro controller unit}
\acro{gtsam}[GTSAM]{Georgia Tech Smoothing and Mapping library}
\acro{swe}[SWE]{Sliding Window Estimator}
\acro{uav}[UAV]{unmanned aerial vehicle}
\acro{dso}[DSO]{Direct Sparse Odometry}
\acro{VersaVIS}[VersaVIS]{Open Versatile Multi-Camera Visual-Inertial Sensor Suite}
\acro{pcb}[PCB]{printed circuit board}
\acro{tof}[ToF]{time of flight}
\acro{rgbdi}[RGB-D-I]{Color-Depth-Inertial}
\acro{lidar}[LiDAR]{light detection and ranging sensor}
\acro{mcu}[MCU]{microcontroller unit}
\acro{usb}[USB]{universal serial bus}
\acro{lidar}[LiDAR]{Light Detection and Ranging sensor}
\acro{ros}[ROS]{Robot Operating System}
\acro{spi}[SPI]{Serial Peripheral Interface}
\acro{ae}[AE]{auto-exposure}
\acro{ros}[ROS]{Robot Operating System}
\acro{pps}[PPS]{Pulse per second}
\acro{mav}[MAV]{micro aerial vehicle}
\acro{i2c}[I$^2$C]{Inter-Integrated Circuit}
\acro{ros}[ROS]{Robot Operating System}
\acro{fov}[FOV]{field of view}
\acro{uart}[UART]{Universal Asynchronous Receiver Transmitter}
\acro{os}[OS]{operating system}
\acro{snr}[SNR]{signal to noise ratio}
\acro{ptp}[PTP]{Precision Time Protocol}
\acro{tsdf}[TSDF]{Truncated Signed Distance Function}
\acro{uav}[UAV]{unmanned aerial vehicle}
\acro{bm}[BM]{block-matching}
\acro{led}[LED]{light emitting diode}
\acro{fir}[FIR]{finite impulse response}
\end{acronym}

\end{document}